\documentclass[pdflatex,iicol]{sn-jnl}

\usepackage{graphicx}%
\usepackage{multirow}%
\usepackage{amsmath,amssymb,amsfonts}%
\usepackage{amsthm}%
\usepackage{mathrsfs}%
\usepackage[title]{appendix}%
\usepackage{xcolor}%
\usepackage{textcomp}%
\usepackage{manyfoot}%
\usepackage{booktabs}%
\usepackage{algorithm}%
\usepackage{algorithmicx}%
\usepackage{algpseudocode}%
\usepackage{listings}%
\usepackage{caption}
\usepackage{subcaption}
\usepackage{colortbl}
\usepackage[table]{xcolor}
\usepackage{natbib}
\usepackage{float}

\theoremstyle{thmstyleone}%
\theoremstyle{thmstyletwo}%

\theoremstyle{thmstylethree}%
\begin{document}

\flushbottom

\title[Article Title]{Offline-Online Curriculum RL for Multimodal Reasoning}


\author[1]{\fnm{Wendi} \sur{Deng}}\email{dengwendi02@bupt.edu.cn}

\author[1]{\fnm{Hang} \sur{Du}}\email{7597892@bupt.edu.cn}

\author*[1,2]{\fnm{Guoshun} \sur{Nan}}\email{nanguo2021@bupt.edu.cn}
\author[1]{\fnm{Haokun} \sur{Tian}}\email{thksang@bupt.edu.cn}
\author[1]{\fnm{Jiaqi} \sur{Yu}}\email{yjiaqi@bupt.edu.cn}
\author[1]{\fnm{Xinlei} \sur{Cao}}\email{2023211537cxl@bupt.edu.cn}
\author[1]{\fnm{Jiale} \sur{Li}}\email{lijiale2023@bupt.edu.cn}
\author[3]{\fnm{Jingfeng} \sur{Chen}}\email{1357696493steven@gmail.com}
\author[1]{\fnm{Ling} \sur{Deng}}\email{2024211778@bupt.cn}
\author[6]{\fnm{Ting} \sur{Li}}\email{liting.sc@chinatelecom.cn}
\author[7]{\fnm{Hao} \sur{Yang}}\email{hyang@csust.edu.cn}
\author[4]{\fnm{Jun} \sur{Liu}}\email{J.liu81@lancaster.ac.uk}
\author[5]{\fnm{Xudong} \sur{Jiang}}\email{exdjiang@ntu.edu.sg}
\author[5]{\fnm{Sicong} \sur{Leng}}\email{Lengsicong@gmail.com}

\affil*[1]{\orgname{Beijing University of Posts and Telecommunications}, \orgaddress{\city{Beijing}, \postcode{100876}, \country{China}}}

\affil[2]{\orgname{BUPT Shenzhen Institute}, \orgaddress{\city{ShenZhen}, \postcode{518038}, \country{China}}}

\affil[3]{\orgname{Carnegie Mellon University}, \orgaddress{\city{Pittsburgh}, \country{USA}}}

\affil[4]{\orgname{Lancaster University}, \orgaddress{\city{Lancaster}, \country{UK}}}

\affil[5]{\orgname{Nanyang Technological University}, \orgaddress{\city{Singapore}, \country{Singapore}}}

\affil[6]{\orgname{China Telecom Corporation Limited Sichuan Branch}, \orgaddress{\city{Chengdu}, \state{Sichuan}, \country{China}}}

\affil[7]{\orgname{Changsha University of Science \& Technology}, \orgaddress{\city{Changsha}, \state{Hunan}, \country{China}}}

\vspace{25pt}
\abstract{
Multimodal large language models exhibit capabilities on reasoning tasks, yet often produce flawed intermediate steps while yielding correct final answers. This behavior undermines interpretability and reliability, suggesting reliance on spurious shortcuts rather than faithful reasoning. Although efforts have explored step-level supervision, distinguishing decisive steps from redundant ones remains challenging. We propose O²-CritiCuRL, a novel curriculum reinforcement learning framework that introduces critical-step awareness through an iterative offline–online paradigm. In the offline stage, O²-CritiCuRL conducts multi-rollout analysis over step-annotated trajectories to estimate step-level importance, allowing the framework to distill critical reasoning steps and filter out redundant ones. In the online stage, we employ a progressive step-level reinforcement learning strategy, where truncated chains guide the model to infer missing steps and refine its reasoning, thereby sharpening its focus on critical steps and overcoming the limitations of static supervision. Extensive experiments on multimodal reasoning benchmarks show that our method achieves state-of-the-art performance while delivering superior training and inference efficiency. Code is available at https://github.com/kk0013/CritiCuRL.}
\vspace{25cm}

\maketitle
\section{Introduction}
Recent advances in large language models (LLMs) have demonstrated remarkable capabilities in multi-step reasoning across domains such as mathematics, code generation, and scientific discovery~\citep{wang2023scientific,guo2024deepseek,comanici2025gemini}. A common strategy for enhancing these reasoning abilities is curriculum learning, which, inspired by human education, gradually exposes models to increasingly complex examples~\citep{bengio2009curriculum,parashar2025curriculumreinforcement}. By regulating task difficulty, curriculum learning stabilizes optimization and improves generalization, making it an effective paradigm for reasoning-intensive tasks~\citep{zeng2024scaling,el2025competitive,lightman2024letsverifystepbystep}. Reinforcement learning (RL) further strengthens reasoning by optimizing models with outcome- or process-based rewards~\citep{hendrycks2021measuringmathematicalproblemsolving,stiennon2022learningsummarizehumanfeedback,wang2024mathshepherdverifyreinforcellms}. Among existing RL methods, Group Relative Policy Optimization (GRPO) has gained increasing attention due to its efficiency and scalability, encouraging models to explore and refine multiple reasoning paths through ``slow thinking’’\citep{shakya2023reinforcement,comanici2025gemini,guo2025deepseek,park2025multiagent,wang2025offlinereinforcementllmreasoning}. Consequently, integrating curriculum learning with reinforcement learning has emerged as a promising paradigm for improving performance on reasoning-intensive tasks\citep{yuan2025vl,shao2024deepseekmath,yao2023treeofthoughts,xie2024mctspreference}.

\begin{figure}[htbp]
\centering
\includegraphics[width=\columnwidth]{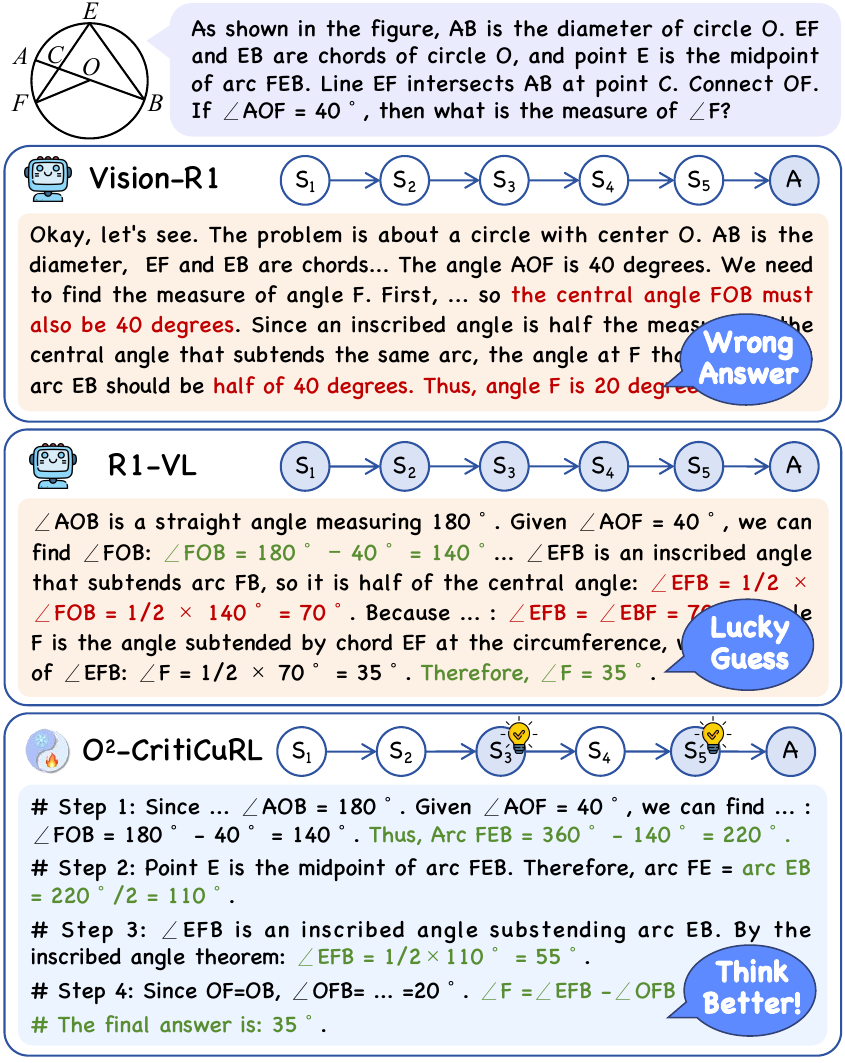}
\caption{\textbf{Comparison of models on mathematical reasoning problems.} Vision-R1 with answer-level supervision fails to derive the correct answer because of incorrect intermediate assumptions. R1-VL supervises every reasoning step and may reach the correct final answer, but still follows a partially flawed reasoning path. In contrast, our O$^2$-CritiCuRL identifies and emphasizes critical reasoning steps, enabling the model to learn accurate, concise, and trustworthy reasoning paths.}
\label{fig:intro}
\vspace{-17pt}
\end{figure}

Nevertheless, strong answer accuracy does not necessarily imply a reliable reasoning process. As illustrated in Figure~\ref{fig:intro}, Multimodal Large Language Models (MLLMs) often produce incorrect, redundant, or unnecessarily complicated intermediate reasoning while still arriving at the correct final answer~\citep{ling2023deductive,lightman2023letsverifystepstep}. For example, a model may introduce an invalid assumption early in a mathematical derivation, compensate for it through subsequent calculations, and eventually obtain the correct result. Although the final answer is correct, the underlying reasoning trajectory remains logically flawed and difficult to trust. Such behavior suggests that models may rely on spurious correlations or shortcuts rather than robust logical deduction~\citep{poursabzi2021manipulating,geirhos2020shortcut}. This substantially limits their applicability in high-stakes scenarios that require transparent and verifiable reasoning, including education, scientific research, and legal or medical decision support~\citep{arrieta2020explainable,benda2022trust,barata2023reinforcement,jayaraman2024primer}. In this sense, the phenomenon resembles reward hacking: instead of learning valid reasoning procedures, the model exploits answer-level supervision that rewards correct outcomes without constraining the intermediate reasoning process~\citep{amodei2016concrete,stiglic2020interpretability}.

A natural solution is to incorporate step-level supervision, which constrains intermediate reasoning by aligning model behavior with annotated reasoning trajectories~\citep{cheng2024chainlm,luo2024improve}. However, not all reasoning steps contribute equally to solving a problem. As illustrated in Figure~\ref{fig:intro}, the example shown contains five explicit reasoning steps, while more challenging mathematical or scientific problems may involve hundreds of intermediate steps, including indispensable deductions, repeated verifications, redundant explanations, and even erroneous detours~\citep{qu2025survey,lu2025prolonged}. Among these steps, only a small subset may be decisive for reaching the correct solution. Treating every step equally not only incurs substantial annotation and training costs, but also dilutes supervision over the truly critical reasoning steps. Consequently, the model may learn to imitate lengthy trajectories rather than capture the essential logical structure required for robust generalization.

Our empirical analysis reveals that only a small subset of intermediate reasoning steps has a decisive impact on the correctness of the final answer, whereas supervising all steps uniformly often introduces substantial redundancy and may even reinforce uninformative reasoning patterns. 
We further observe that the locations and types of such decisive steps are not static but evolve as the model’s reasoning ability improves (see Section~\ref{sec:discussion} for detailed analysis).
Existing RL methods with process-level rewards offer a promising alternative to trajectory imitation by enabling the model to optimize its intermediate reasoning behavior dynamically~\citep{wang2025towards,chen2025towards}. However, these methods typically depend on additional reward models, incur considerable computational overhead, and remain susceptible to inaccurate reward estimation and reward hacking. More importantly, because the model’s reasoning strategies and failure modes continuously change during training, a fixed curriculum or a set of critical steps identified before training may gradually become misaligned with the model’s current capability.

These limitations give rise to two fundamental issues: \textcircled{1} accurately identifying the small subset of reasoning steps that truly determines the correctness of a solution, and \textcircled{2} dynamically updating the corresponding critical-step supervision as the model evolves throughout training.
Motivated by these observations, our key insight is that effective reasoning supervision should concentrate on the dynamically evolving decisive steps of a reasoning trajectory, rather than treating all intermediate steps indiscriminately. By identifying, emphasizing, and continuously adapting supervision over these critical steps, the model can learn more concise, reliable, and logically consistent reasoning patterns while requiring substantially less redundant supervision.

Thus, we propose O$^2$-CritiCuRL, an offline–online curriculum reinforcement learning framework for critical-step-aware reasoning. O$^2$-CritiCuRL combines stable offline curriculum construction with adaptive online reinforcement learning. The offline component accurately identifies the small subset of reasoning steps that truly determine the correctness of a solution from existing trajectories, thereby addressing the issue \textcircled{1}, and organizes training samples according to reasoning difficulty. The online component progressively refines the model using step-level reward signals. By iteratively updating critical-step identification with the improved model, the resulting curriculum remains aligned with the model’s evolving reasoning ability, thereby addressing the issue \textcircled{2}.

In summary, our contributions are threefold:
\begin{itemize}
\item We propose O$^2$-CritiCuRL, a novel critical-step-aware curriculum reinforcement learning framework that identifies the reasoning steps most influential to solution correctness and constructs adaptive curricula around them.
\item We develop an efficient iterative offline–online optimization mechanism that combines stable and efficient offline critical-step identification with adaptive online reinforcement learning, enabling the curriculum to remain aligned with the model’s evolving reasoning ability.
\item We conduct extensive experiments demonstrating that O$^2$-CritiCuRL consistently improves reasoning performance and training efficiency, while producing more concise, reliable, and logically consistent reasoning trajectories with stable optimization.
\end{itemize}

\section{Related Work}

\subsection{Curriculum \& RL Reasoning}
Curriculum learning is a training paradigm that organizes tasks in increasing order of difficulty to promote effective model learning~\citep{bengio2009curriculum，nan2023physical}. 
In the domain of LLMs and VLMs, it has been widely adopted to stabilize optimization and enhance reasoning generalization. 
From the perspective of reward design, existing works can be categorized as result-based or process-based rewards. 
Result-based methods evaluate the final answer correctness as the primary criterion, such as MMR1~\citep{leng2025mmr1},  VLM-R1~\citep{shen2025vlmr1}, Vision-G1~\citep{zha2025vision}, Infi-MMR~\citep{liu2025infimmr}, ASTRO~\citep{kim2025astro}, and JT-Math~\citep{hao2025jt}. 
In contrast, process-based reward approaches explicitly assess intermediate steps, as in R-PRM~\citep{she2025r}, VisualPRM~\citep{wang2025visualprm}. 
These methods provide finer-grained supervision signals but still face challenges such as high reward modeling costs.  
From the perspective of training phase design, works like Curr-ReFT~\citep{deng2025boosting} and PCuRL~\citep{yuan2025vl} introduce explicit progressive stages, while E2D~\citep{parashar2025curriculum}, and GHPO~\citep{liu2025ghpo} employ dynamic scheduling functions or adapt between imitation and reinforcement learning. 
Although effective for structuring learning, these methods typically rely on static difficulty labels or fixed schedules and overlook the importance of intermediate reasoning steps.
Our work introduce a curriculum reinforcement learning framework that explicitly integrates step-level supervision with critical-step awareness.

\subsection{Supervision for Reasoning}
Traditional answer-only supervision enables models to bypass reasoning trajectories as long as the final answer is correct, which undermines interpretability and reliability~\citep{lightman2023let, du2024exploring}. 
To mitigate this issue, recent efforts incorporate intermediate reasoning rewards, such as R-PRM~\citep{she2025r}, VisualPRM~\citep{wang2025visualprm}, and StepGRPO~\citep{zhang2025r1}, which evaluate the quality of intermediate steps. 
Although these methods provide transparent supervision, they tend to treat all steps equally, ignoring the fact that many reasoning trajectories contain redundant or low-impact steps~\citep{qu2025survey,arora2025training}. 
Several approaches attempt to address this issue through data selection or trajectory restructure. 
For example, DUMP~\citep{wang2025dump}, Writing-RL~\citep{lei2025writing}, and Vision-G1~\citep{zha2025vision} prioritize high-value samples via bandit mechanisms or influence functions, while EduFlow~\citep{zhu2025eduflow} and VersaPRM~\citep{zeng2025versaprm} leverage large-scale preference annotations. 
However, these strategies remain limited in their ability to dynamically pinpoint which reasoning steps are critical for the problems.  
Our work differs from prior studies by introducing an offline–online framework that detects and iteratively refines decisive reasoning steps, enabling improved accuracy, interpretability, and training efficiency.

\section{Theoretical Proofs}
\label{theoretical proofs}
\textbf{Assumption}: We define a critical step as a reasoning step that induces a substantial shift in the model's answer distribution toward the correct answer, and consider the model's state as a conditional distribution over possible answers given the current reasoning prefix. A reasoning step is regarded as critical if it significantly reduces the model's uncertainty or redirects probability mass from incorrect regions toward the correct answer. 
\noindent
\textbf{Definition}: We begin by defining the probabilistic state space of the reasoning process. 
Let $\Omega$ be the set of all possible answers. 
The model's reasoning state is represented by a probability distribution over $\Omega$.
$P_0=P(\cdot\mid q)$ denotes the model's initial answer distribution given the question without any reasoning steps, while $P_i$ denotes the conditional distribution given steps $S_{1:i}$.
\begin{equation}
  P_{i}=P(\cdot\mid q, S_{1:i})
\end{equation}
To measure the uncertainty of the answer space, we introduce the Shannon Entropy, and the contribution of the i-th step can be quantified as the entropy reduction in $\Delta H_i$:
\begin{equation}
\resizebox{\linewidth}{!}{$
\begin{gathered}
  H(P)=-\sum_{a\in\Omega}P(a)\log P(a),\;\Delta H_i=H(P_{i-1})-H(P_i)
\end{gathered}
$}
\end{equation}
Within our formulation, the reduction in uncertainty of the conditional distribution directly manifests as compression of the probability space, effectively eliminating a substantial portion of incorrect answers. (i.e., entropy reduction). 
The degree and efficiency of probability distribution changes triggered by each reasoning step are central to measuring its criticality. 
The \textbf{relative entropy} (Kullback-Leibler Divergence) naturally quantifies this distribution change, making it our information-theoretic measure for the criticality of step $S_i$:
\begin{equation}
\mathcal{K}_i = D_{\text{KL}}(P_i \parallel P_{i-1}) = \sum_{\omega \in \Omega} P_i(\omega) \log \frac{P_i(\omega)}{P_{i-1}(\omega)}
\end{equation}
\noindent\textbf{Step 1: Decomposition of Relative Entropy.}
We decompose the relative entropy into contributions from the correct answer $\omega^*$ and the space of incorrect answers $\omega$, the first term is the contribution from correct answer probability increase, and the second term reflects the distributional change within the incorrect-answer subspace:
\begin{equation}
\begin{aligned}
    \mathcal{K}_i = P_i(\omega^*) \log \frac{P_i(\omega^*)}{P_{i-1}(\omega^*)} + \sum_{\omega \neq \omega^*} P_i(\omega) \log \frac{P_i(\omega)}{P_{i-1}(\omega)}
\end{aligned}
\end{equation}

\noindent\textbf{Step 2: Coarse-Grained Decomposition.}
To characterize step-level distribution changes in multimodal reasoning, we decompose the answer-space shift into a coarse-grained correctness term and a residual term within the incorrect-answer subspace. The former measures how much probability mass is shifted toward the correct answer, while the latter captures the redistribution among incorrect answers, including possible semantic clusters caused by visual perception errors, reasoning shortcuts, or similar answer candidates.
Let $p_i=P_i(\omega^*)$ denote the probability assigned to the correct answer at step $S_i$. For the incorrect-answer subspace, we define the normalized conditional distribution:
$
Q_i(\omega)=\frac{P_i(\omega)}{1-p_i}, \quad \omega\neq \omega^* .
$
Accordingly, for each incorrect answer $\omega\neq\omega^*$, we have
$
P_i(\omega)=(1-p_i)Q_i(\omega).
$
Substituting this decomposition into the KL divergence gives
\begin{equation}
\begin{aligned}
K_i
=
p_i\log\frac{p_i}{p_{i-1}}
+
(1-p_i)\log\frac{1-p_i}{1-p_{i-1}} \\
+
(1-p_i)D_{\mathrm{KL}}(Q_i\|Q_{i-1}).
\end{aligned}
\end{equation}
\textbf{Step 3: Computable Correctness-Level Approximation.}
In practice, our rollout protocol directly estimates the correctness-level transition after each reasoning prefix. Therefore, we adopt the coarse-grained correctness projection, which groups the answer space into two events: correct and incorrect. This yields a tractable Bernoulli KL term:
\begin{equation}
    K_i^{\mathrm{cg}}=
    p_i\log\frac{p_i}{p_{i-1}}
    +
    (1-p_i)\log\frac{1-p_i}{1-p_{i-1}}     
\end{equation}

\noindent\textbf{Step 4: Criticality Metric Formula.}
Since $D_{\mathrm{KL}}(Q_i\|Q_{i-1})\geq 0$, the coarse-grained score $K_i^{\mathrm{cg}}$ provides a conservative lower-bound approximation of the full answer-space KL divergence. Based on the above analysis, we use the coarse-grained correctness-level KL as the practical criticality metric. For simplicity, we denote $K_i^{\mathrm{cg}}$ as $K_i$ in the following sections:
\begin{equation}
\boxed{\displaystyle
\mathcal{K}_i
= p_i \log\frac{p_i}{p_{i-1}}
+ (1-p_i)\log\frac{1-p_i}{1-p_{i-1}}
}
\label{K-reward}
\end{equation}
This metric is fully characterized by the model's probability of producing the correct answer before and after introducing step $S_i$. The previous uniform compression case can be viewed as a special case where $Q_i=Q_{i-1}$, meaning that the relative distribution inside the incorrect-answer subspace remains unchanged. 

\paragraph{Boundary Behavior and Robustness.} 
The criticality measure $\mathcal{K}i$ has intuitive boundary behavior: it grows large when a step increases the probability from near zero to a meaningful value, and it approaches zero as $p_{i-1}\to 1$, where no further entropy reduction is possible.
In practice, the model rarely reaches such extremes, as neither infinitely informative nor completely negligible steps occur under finite model capacity.
Furthermore, the logarithmic form naturally suppresses minor fluctuations: when $p_i \approx p_{i-1}$ and both are moderate, $\mathcal{K}i$ becomes much smaller than $|p_i - p_{i-1}|$, reducing noise-induced false positives.

\section{Method}
\begin{figure*}[htbp]
    \centering
    \includegraphics[width=\textwidth]{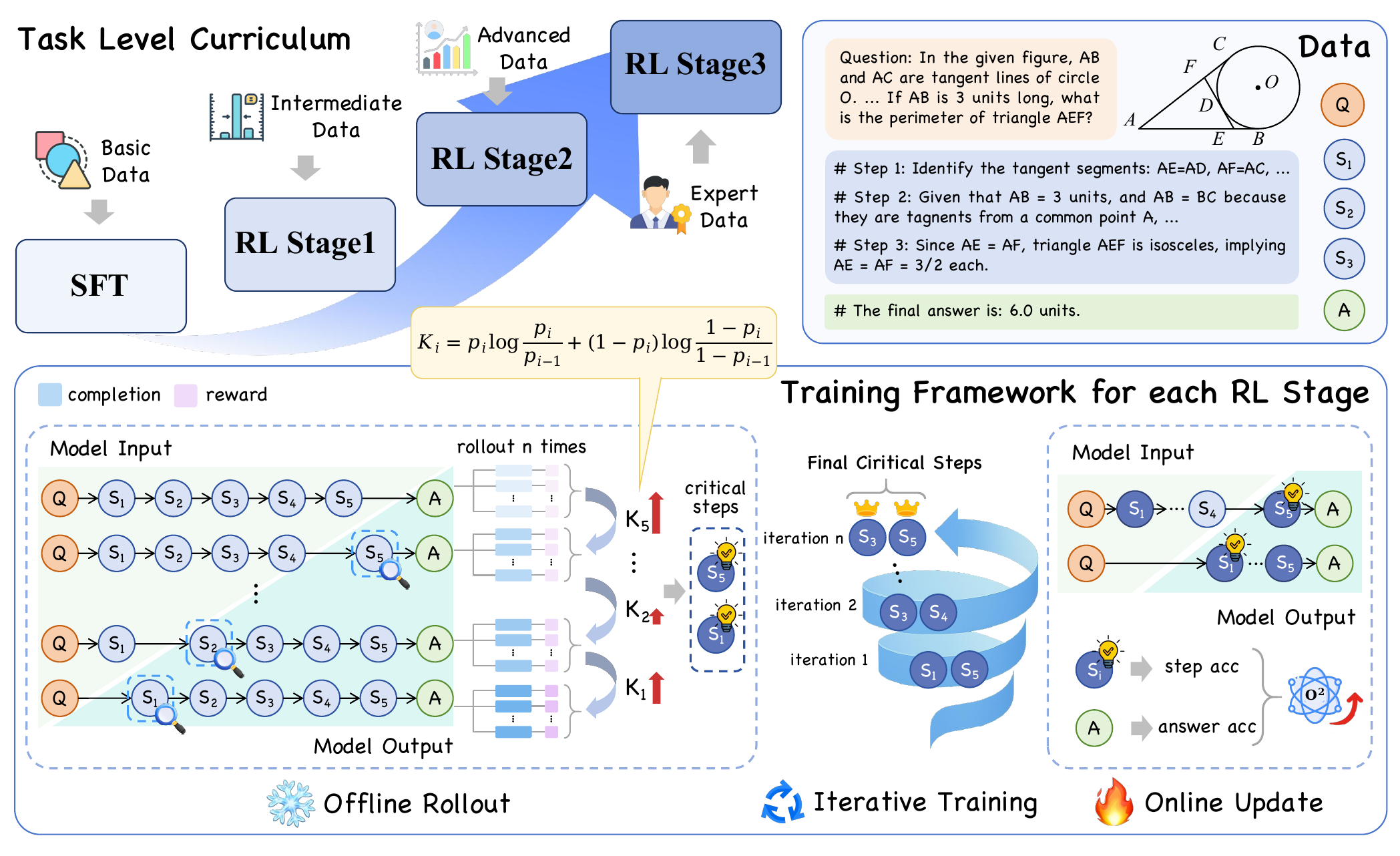}
    \caption{Our O²-CritiCuRL framework conducts a two-level curriculum. (1) Task-level curriculum: data are organized by increasing difficulty, progressing from SFT to RL-Stage1/2/3. (2) Critical-step curriculum: each RL stage alternates between offline rollout and online update. The offline phase uses a KL-based score $K_i$ to identify decisive steps. The online phase then optimizes the model with step-wise rewards guided by these steps. This iterative loop continuously refines critical steps and strengthens reasoning.} \label{fig:method_main}
\vspace{-17pt}
\end{figure*}

In this section, we illustrate our proposed framework O²-CritiCuRL in detail.
The method begins with supervised fine-tuning on step-annotated data to provide a cold-start initialization.
To estimate difficulty, each sample is evaluated by GPT over multiple trials, and its answer accuracy is recorded.
Based on GPT correctness and the number of reasoning steps, all samples are grouped into difficulty tiers, and each training stage focuses on one tier, enabling a curriculum that gradually strengthens the model from easy to hard tasks.
Building on this foundation, we adopt an iterative optimization strategy in which offline and online phases alternate.
In the offline phase, key reasoning steps are extracted from existing trajectories and organized into curriculum signals. 
In the online phase, these signals guide reinforcement learning updates to refine the model’s reasoning policy.
The updated model is then returned to the offline phase for refined critical-step identification, forming a closed offline–online loop.
This process repeats until convergence, ensuring both stable critical-step discovery and effective reinforcement learning.

\subsection{Data Processing}
To estimate data difficulty, each question is submitted to GPT for ten independent trials, and the response accuracy across these attempts is recorded.
Based on the correctness rate and the number of reasoning steps required to reach the correct answer, we categorize the samples into four curriculum stages to ensure a progressive training scheme from easy to hard.
Stage $1$ includes samples correctly answered in all ten trials with short reasoning chains ($1–3$ steps); Stage $2$ covers those fully correct but with longer reasoning chains ($\geq$~4 steps); Stage $3$ consists of samples correctly answered in $2–9$ trials; and Stage $4$ contains the most difficult ones, answered correctly in at most one trial.
To mitigate the risk of reward hacking during reinforcement learning, all multiple-choice questions in the original dataset are converted into an open-ended question-answering format.
This stage-wise organization enables the model to be trained under a curriculum learning paradigm, gradually enhancing its reasoning capability from simple to complex problems while maintaining stability and efficiency in optimization.

\subsection{Model Initialization}
We adopt Qwen-2.5-VL~\citep{bai2025qwen25vltechnicalreport} as the backbone model and initiate the training process with a cold-start strategy.
We randomly sample a subset of Mulberry~\citep{wang2023scientific} for SFT stage.
To enable curriculum learning that gradually exposes the model to harder samples, we use the remaining data in our framework and structure training into three stages.
Unlike conventional curriculum learning approaches that primarily adjust task difficulty in a coarse-grained manner, we introduce a two-phase progressive refinement framework in each stage. 
\subsection{Offline Step Distillation} \label{sec:offline-step-distillation}
In the offline phase, the model autonomously identifies the critical reasoning steps as shown in the lower-left panel of Figure \ref{fig:method_main}. 

For each instance, the segmented reasoning steps are concatenated with the original question sequentially. 
Each step is appended to the question one by one, forming inputs of the model, until the complete chain of reasoning is reconstructed:
\begin{equation}
\mathcal{I}
= \Bigl\{
    q,\;
    q + s_1,\;
    q + s_1 + s_2,\;
    \ldots,\;
    q + \sum_{t=1}^{T} s_t
  \Bigr\},
\label{eq:input_sequence}
\end{equation}
We sequentially feed the constructed inputs into the model and evaluate the reward based on whether the predicted answer matches the ground truth. 
For each input, we perform $n$ independent rollouts, obtaining $n$ reward samples. 
These samples form an empirical distribution that approximates the model’s behavior over the answer space for the given input.
Following the practical refinement of Equation \ref{K-reward}, we introduce several adjustments to ensure stability under extreme probability cases. 
First, both $p_i$ and $p_{i-1}$ are clipped into the interval $[\varepsilon, 1-\varepsilon]$ to avoid degenerate values at $0$ or $1$.
Second, a small lower bound is imposed when $p_i$ falls below a threshold (e.g., $0.01$), preventing disproportionately large contributions to the metric.

Under these refinements, we leverage the stabilized probabilities $m_i$ and $m_{i-1}$ to compute the criticality score $K$ between adjacent steps as following:
\begin{equation}
K_{i} = m_i \log \frac{m_i}{m_{i-1}} + (1-m_i) \log \frac{1-m_i}{1-m_{i-1}}
\end{equation}
where $m_i$ and $m_{i-1}$ denote the probability values after the adjustments.
Based on the ranking of computed criticality scores, we select the $top_{k}$ steps with the highest scores as the identified critical steps.
Specifically, $k$ is defined as
\begin{equation}
    k = \max\left(1, \left\lfloor \frac{T}{3} \right\rfloor\right)
\end{equation}
where $T$ denotes the total number of reasoning steps. This adaptive strategy allows longer trajectories to retain more critical steps while preventing short trajectories from introducing redundant supervision.

\subsection{Online Step Optimization} 
As illustrated in the lower-right panel of Figure~\ref{fig:method_main}, we construct paired training samples for reinforcement learning. 
Each pair consists of a truncated input sequence and its corresponding target, where critical steps identified during offline distillation are explicitly marked. 
This formulation enables the model to receive supervision not only from the final answer but also from intermediate reasoning steps, thereby emphasizing the contribution of critical steps during optimization.
To encourage the model to derive correct answers by accurately reasoning through critical steps, we incorporate a reward component that explicitly reflects step-wise correctness. 
The reward function is defined in Equation~\ref{eq:score}, where $\alpha$, $\beta$, and $\gamma$ maintain the balance among different rewards.

\begin{equation}
\mathscr{S}_i =\alpha R_a + \beta R_f + \gamma (R_t + R_s)
\label{eq:score}
\end{equation}
$R_{\text{f}}$ and $R_{\text{t}}$ evaluates the step format and the answer format, $R_{\text{s}}$ and $R_{\text{a}}$ measure the correctness of reasoning steps and the final answer. 
The online strategy design encourages the model to internalize both the accuracy of the final answer and the soundness of intermediate reasoning steps, thereby fostering the ability to derive correct answers through coherent and well-structured reasoning. \\

The existing GRPO~\citep{shao2024deepseekmath, cao2025advancing} framework is adopted for the reinforcement learning paradigm. To further constrain optimization to critical reasoning steps, 
We perform \emph{targeted rollouts} for prefixes leading to critical steps, reducing computation and sharpening the training focus on decisive reasoning steps. 
As shown in Equation~\ref{eq12}, we let $\mathcal{K}_i = \{k_1, k_2, \ldots, k_m\}$ denote the indices of critical steps in the $i$-th reasoning trajectory $\{s_1, s_2, \ldots, s_{|o_i|}\}$. 
For each $k_j \in \mathcal{K}_i$, a rollout sample is constructed using the prefix $s_{<k_j}$ as input and the subsequence $s_{k_j:|o_i|}$ (i.e., the remaining reasoning steps and final answer $A$) 
as the target. 
This design ensures that policy optimization is conducted only if the next predicted step corresponds to a critical transition, thereby emphasizing information gain at decisive reasoning junctures. 
\begin{equation}
q + s_1 + \dots + s_{k_i-1} \longrightarrow s_{k_i} + \dots + s_n + \mathbf{A}
\label{eq12}
\end{equation}


\subsection{Offline-Online Iteration}

As training progresses, the model’s ability to identify critical steps improves, yielding step selections that better align with its current reasoning capabilities and exert greater influence on its decision process.
Importantly, the set of steps regarded as critical is not static, as previously difficult steps may no longer be bottlenecks when the model improves, while new key steps may become more important for reasoning success.
This dynamic evolution provides learning signals that are better aligned with the model’s evolving reasoning capabilities.
For each epoch, critical steps within all samples are first identified through \textit{Offline Step Distillation}. 
The samples are then reorganized during the \textit{Online Step Optimization} and used to update the model, marking the completion of one iteration. 
In the subsequent iteration, the updated model is applied to the next epoch, where \textit{Offline Step Distillation} and \textit{Online Step Optimization} are repeated.
Through this iterative cycle, the model progressively improves both its capacity to recognize critical steps and its accuracy in reasoning. 
The iterative process proceeds until model's reward signal reaches a stable level, at which point the reinforcement learning stage is considered converged and training advances to the subsequent phase.

\section{Experiments}

\begin{table*}[t]
\renewcommand{\arraystretch}{1.2}
\centering
\caption{Comparisons on multiple multimodal reasoning benchmarks.}
\setlength{\tabcolsep}{1.5pt}
\resizebox{\textwidth}{!}{
\begin{tabular}{lcccccccccc}
\toprule
\multirow{2}{*}{\textbf{Model}} & \multirow{2}{*}{\textbf{Size}}
& \multicolumn{5}{c}{\textbf{Mathematics}}
& \multicolumn{2}{c}{\textbf{Science}}
& \multicolumn{2}{c}{\textbf{General}} \\
\cmidrule(lr){3-7}\cmidrule(lr){8-9}\cmidrule(lr){10-11}
& & MathVista & MathVision & MathVerse & ChartQA & LogicVista
& ScienceQA & MMMU & MMStar & MME$_{sum}$ \\
\midrule
\rowcolor{gray!10} \multicolumn{11}{c}{\textbf{General-Purpose Models}} \\
\midrule
Qwen2.5-VL~\citep{bai2025qwen25vltechnicalreport} & 7B & 61.9 & 21.9 & 40.1 & 80.7 & 39.1 & 87.2 & 50.9 & 61.0 & 2303 \\
InternVL2.5~\citep{chen2024expanding} & 8B & 57.8 & 18.4 & 40.0 & 75.4 & 36.8 & 88.3 & 43.1 & 62.0 & 2218 \\
InternVL3~\citep{zhu2025internvl3exploringadvancedtraining} & 8B & 62.6 & 28.6 & 39.8 & 81.8 & 39.3 & 89.1 & 50.8 & 62.5 & 2424 \\
LLaVA-OneVision~\citep{li2025llavaonevision} & 7B & 53.3 & 15.3 & 33.6 & 66.1 & 30.6 & 80.5 & 41.6 & 60.7 & 1487 \\
DeepSeek-VL~\citep{lu2024deepseekvl} & 7B & 36.1 & 13.7 & 26.2 & 59.1 & - & 88.6 & 36.6 & 37.1 & 1790 \\
MiniCPM-o-2.6~\citep{openbmb2025minicpmo} & 8B & 69.0 & 21.6 & 35.0 & 83.6 & - & - & 51.7 & 62.6 & 2328 \\
\midrule
\rowcolor{gray!10} \multicolumn{11}{c}{\textbf{Reasoning-Oriented Models}} \\
\midrule
LLaVA-CoT~\citep{Xu_2025_ICCV} & 11B & 54.8 & 20.0 & 33.9 & 78.9 & - & - & 48.9 & 57.6 & 2137 \\
LLaVA-Reasoner~\citep{zhang-etal-2025-improve} & 8B & 50.6 & 22.3 & 34.1 & 83.0 & - & 86.8 & 40.0 & 54.0 & - \\
Insight-V~\citep{dong2024insight} & 8B & 59.9 & 23.2 & 33.0 & 81.5 & - & 85.3 & 48.0 & 60.4 & 1869 \\
MM-Eureka~\citep{meng2025mmeureka} & 8B & 68.3 & 27.4 & 40.2 & 83.0 & 40.9 & 86.4 & 51.2 & 63.7 & 2286\\
OpenVLThinker~\citep{deng2025openvlthinker} & 7B & 67.4 & 24.0 & 42.1 & 82.3 & 40.5 & 87.4 & 52.3 & 59.1 & 2111 \\
R1-VL~\citep{zhang2025r1vllearningreasonmultimodal} & 7B & 63.2 & 28.2 & 37.6 & 79.9 & 35.9 & 84.5 & 38.1 & 61.0 & 2258 \\
Vision-R1~\citep{huang2025visionr1incentivizingreasoningcapability} & 7B & 69.2 & 26.9 &\textbf{42.9} & 83.1 & 41.2 & 87.9 & 35.9 & 62.1 & 2268 \\
Mulberry~\citep{yao2024mulberryempoweringmllmo1like} & 7B & 59.7 & 26.0 & 41.6 & 80.3 & 34.2 & 85.2 & 39.7 & 59.5 & 2278 \\
R1-Onevision~\citep{yang2025r1onevisionadvancinggeneralizedmultimodal} & 7B & 60.5 & 28.9 & 40.0 & 72.2 & 39.6 & 87.7 & 45.4 & 61.8 & 2342 \\
\midrule
\rowcolor{blue!10}
\textbf{O$^2$-CritiCuRL (Ours)} & 7B & \textbf{69.8} & \textbf{29.3} & 42.1 & \textbf{84.2} & \textbf{41.7} & \textbf{89.3} & \textbf{53.2} & \textbf{64.4} & \textbf{2376} \\
\bottomrule
\end{tabular}
}
\label{tab:vlm-results}
\end{table*}

\begin{figure*}
    \centering
    \vspace{-12pt}
    \includegraphics[width=1.0\linewidth]{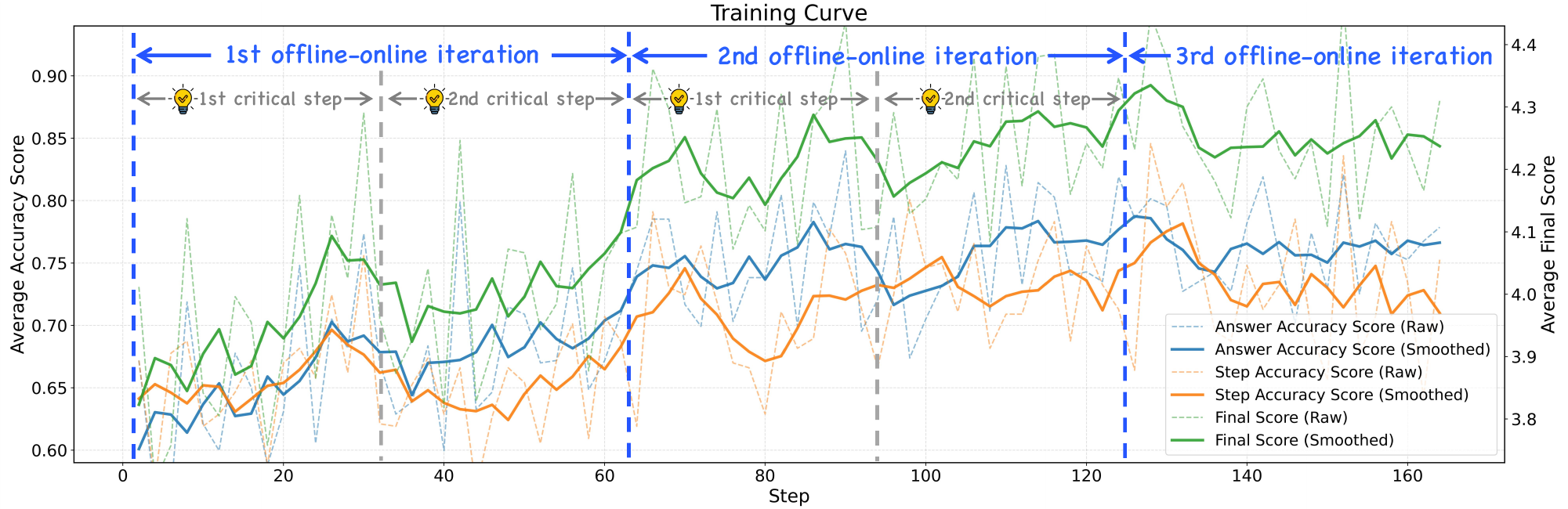}
    \caption{Training curves of step-wise and answer-level rewards.}
    \label{fig:training_curves}
\vspace{-10pt}
\end{figure*}

\subsection{Implementation Details} 
\subsubsection{Evaluation}
We adopt a hierarchical soft matching framework for both final answers and critical reasoning steps. 
For answers, we first normalize predictions and references with LaTeX processing and unit unification. 
They are compared using layered strategies, including exact, equality-based soft, numerical, and canonical containment matching.
Multiple candidates are allowed only on the reference side to capture annotation diversity, while enumerated predictions are disallowed to prevent guessing. 
For reasoning steps, reference keywords are extracted with Qwen3-Max~\citep{qwen3max}, and outputs are segmented into ``\# Step k:'' blocks. 
A strict–soft matching scheme is applied, relying on LaTeX normalization for strict matching and falling back to the answer-level matcher otherwise. 
Enumeration suppression, dynamic length control, and step count limits further ensure precise and efficient detection of critical conclusions.  

\subsubsection{Training Details} 
We randomly sample $187$k examples from Mulberry~\citep{wang2023scientific} for SFT stage and $10$k examples for the Offline–Online RL stages.
All experiments are conducted on $8 \times$ A800 GPUs. Training with $3.3$k questions per stage for one epoch takes about $14$ hours. 
We adopt \texttt{Qwen}\citep{bai2025qwen25vltechnicalreport} as the backbone. 
In the SFT stage, we train for $3$ epochs with a batch size of $32$ and learning rate $2\mathrm{e}{-6}$. 
In the RL stage, the learning rate is set to $1\mathrm{e}{-6}$, and we experiment with subsets of $2.5$k samples trained for a single epoch. 
Reward weights are set to $\alpha=2$, $\beta=1$, and $\gamma=1$.  

\subsection{Main Results}
We evaluate the proposed O²-CritiCuRL on several benchmarks. 
As shown in Table~\ref{tab:vlm-results}, MathVista~\citep{lu2024mathvistaevaluatingmathematicalreasoning}, DynaMath~\citep{zou2025dynamathdynamicvisualbenchmark}, MathVision~\citep{wang2024measuringmultimodalmathematicalreasoning}, and MathVerse~\citep{zhang2024mathversedoesmultimodalllm} assesses visual mathematical reasoning across charts, diagrams, and images.
ChartQA~\citep{chartqa} is a benchmark for question answering about charts with visual and logical reasoning.
O²-CritiCuRL achieves the best results on MathVista, MathVision, ChartQA, and DynaMath, and obtains competitive performance on MathVerse comparable to leading models.
These results demonstrate that our offline–online critical-step curriculum substantially enhances multimodal mathematical reasoning, consistently outperforming strong baselines across diverse task settings.
We also observe that our method achieves comparable results with the strong baseline Vision-R1 on the MathVerse, while performs much better on other ones. 
The underlying reason is that MathVerse includes a non-negligible subset of purely visual problems whose solutions hinge on detailed visual perception~\citep{liu2026tlight} rather than step-wise reasoning, thereby reducing the relative advantage provided by our critical-step optimization.

\subsubsection{Training Stability} 
Figure~\ref{fig:training_curves} shows that although the raw answer-level reward, step-wise reward, and final score exhibit noticeable stochastic fluctuations, their smoothed curves all maintain stable upward trends throughout a full RL training run. Specifically, both the answer-accuracy score and the step-accuracy score improve steadily as training proceeds, and the final score also increases consistently, indicating that all three components are optimized in a coordinated and stable manner rather than improving at the expense of one another. This observation suggests that our training framework provides reliable supervision for both intermediate reasoning quality and final answer correctness, leading to stable end-to-end optimization.

More importantly, the reward continues to improve across offline–online iterations. The second iteration further increases the reward after the improvement achieved in the first iteration, which shows the value of iterative key-step identification and online policy refinement. Within each iteration, each identified key step contributes to reward improvement, with the first key step usually bringing the largest gain. 
This result supports our motivation that key steps have unequal effects on model optimization, and that prioritizing the informative step can lead to the greatest benefit.

\subsubsection{Evaluation of Critical-Step Identification}
To evaluate whether O$^{2}$-CritiCuRL identifies meaningful critical steps, we conduct a model-conditioned human evaluation on 500 questions sampled from MathVista, MathVision, ScienceQA, and MME. Instead of assuming fixed human-defined gold steps, annotators judge whether the selected steps reflect reasonable reasoning bottlenecks under the current model capability, given the question, image, reasoning trajectory, selected steps across offline--online iterations, and model behaviors before and after these steps.
We adopt three metrics. First, \textbf{Model-Conditioned Critical Alignment (MCCA)} measures the agreement between method-selected steps and human-judged model-conditioned critical steps. Let $S_i^{(t)}$ be the steps selected for sample $i$ at iteration $t$, and $H_i^{(t)}$ be the corresponding human-judged critical steps. We define
$
\text{MCCA}^{(t)}=
\frac{2\cdot \text{MCP}^{(t)}\cdot \text{MCR}^{(t)}}
{\text{MCP}^{(t)}+\text{MCR}^{(t)}},
$
where $\text{MCP}^{(t)}$ and $\text{MCR}^{(t)}$ denote the averaged precision and recall between $S_i^{(t)}$ and $H_i^{(t)}$. Second, \textbf{Dynamic Shift Validity (DSV)} evaluates whether critical-step changes across iterations are meaningful:
$
\text{DSV}=\frac{N_{\text{valid}}}{N_{\text{shift}}},
$
where $N_{\text{valid}}$ denotes the number of valid critical-step shifts, and $N_{\text{shift}}$ denotes the total number of observed shifts. Finally, \textbf{Model-Conditioned Critical Deletion Impact (MCDI)} measures the functional necessity of the selected steps:
$
\text{MCDI}^{(t)} = I_{\text{crit}}^{(t)} - I_{\text{random}}^{(t)},
$
where $I_{\text{crit}}^{(t)}$ and $I_{\text{random}}^{(t)}$ denote the invalidity rates after removing the selected critical steps and other random steps, respectively. A larger MCDI indicates that the selected steps are more essential for preserving the reasoning process. As shown in Table \ref{tab:critical_step_evaluation}, O$^{2}$-CritiCuRL achieves the best results across all three metrics, indicating that our KL-based critical-step identification can more effectively capture dynamic and model-dependent reasoning bottlenecks.

\begin{table}[htbp]
\centering
\small
\setlength{\tabcolsep}{1.5pt}
\renewcommand{\arraystretch}{1.2}
\resizebox{\columnwidth}{!}{
\begin{tabular}{lcccc}
\toprule
\textbf{Method} & \textbf{MCCA(\%)} & \textbf{DSV(\%)} & \textbf{MCDI(\%)} & \textbf{Average(\%)}\\
\midrule
$\Delta p$ & 63.7 & 66.2 & 58.0 & 62.6 \\
Normalized $\Delta p$ & 67.5 & 61.4 & 63.3 & 64.1 \\
\textbf{Ours} & \textbf{73.1} & \textbf{81.4} & \textbf{65.7} & \textbf{73.4} \\
\bottomrule
\end{tabular}
}
\caption{Evaluation of critical-step identification.}
\label{tab:critical_step_evaluation}
\vspace{-25pt}
\end{table}

\subsubsection{Reward Hacking Analysis}
To estimate reward-hacking frequency, we manually inspect 500 sampled instances from each of MathVista, ChartQA, ScienceQA, and MME$_{sum}$. Each case is independently reviewed by 10 human experts, with a GPT-based judge used only as an auxiliary reference. A sample is labeled as reward hacking when a strict expert majority agrees that the final answer is correct but the reasoning contains clear errors, inconsistencies, or shortcut behaviors. Under this protocol, our method consistently shows lower reward-hacking rates than strong reasoning baselines. Compared with Vision-R1 and R1-OneVision, it reduces reward hacking by an average of 4.7\%, 3.3\%, 2.5\%, and 3.0\% on MathVista, ChartQA, ScienceQA, and MME$_{sum}$, respectively, suggesting that our offline--online critical-step curriculum improves both task accuracy and reasoning faithfulness.

\subsection{Ablation Study}

\subsubsection{Training Stages}
To validate the necessity of the multi-stage curriculum learning strategy, we conduct experiments to evaluate the performance gains achieved by the model at each stage. 
As shown in Table \ref{tab:stages-results}, accuracy on each benchmark improves at every stage of training. 
SFT brings initial gains over the base model, while RL-Stage1 and RL-Stage2 provide substantial boosts due to the model’s improved ability to leverage the distilled critical steps. 
RL-Stage3 yields an additional improvement, suggesting that iteratively refining critical-step reasoning continues to provide learning signals even after earlier stages have converged.

\newcommand{\bluedelta}[1]{%
  {\footnotesize\textcolor{blue}{\textbf{#1}}}%
}

\begin{table*}[htbp]
\centering
\small
\renewcommand{\arraystretch}{1.0}
\setlength{\tabcolsep}{1.3pt}
\resizebox{\textwidth}{!}{
\begin{tabular}{l|c|cccccccc}
\toprule
\textbf{Models} & \textbf{Average} & \textbf{MathVista} & \textbf{MathVision} & \textbf{MathVerse} & \textbf{ChartQA} & \textbf{ScienceQA} & \textbf{MMMU} & \textbf{MMStar} & \textbf{MME$_{sum}$} \\
\midrule
Base      & 57.5  & 61.9 & 21.9 & 39.1 & 80.7 & 87.2 & 50.9 & 61.0 & 2303 \\
\midrule
+ SFT          & 59.0 & 64.0 & 23.7 & 40.6 & 82.3 & 87.9 & 52.2 & 62.5 & 2334 \\
\midrule
\rowcolor{blue!10}
+ RL-Stage1     & 59.9 & 65.2 & 25.4 & 41.3 & 82.9 & 88.4 & 52.6 & 63.3 & 2350 \\
$\Delta$         & \bluedelta{+0.9} 
                 & \bluedelta{+1.2}
                 & \bluedelta{+1.7}
                 & \bluedelta{+0.7}
                 & \bluedelta{+0.6}
                 & \bluedelta{+0.5}
                 & \bluedelta{+0.4}
                 & \bluedelta{+0.8}
                 & \bluedelta{+16} \\
\midrule
\rowcolor{blue!10}
+ RL-Stage2     & 61.1 & 68.8 & 27.6 & 41.7 & 83.8 & 89.0 & 52.8 & 63.9 & 2368  \\
$\Delta$         & \bluedelta{+1.2}
                 & \bluedelta{+3.6}
                 & \bluedelta{+2.2}
                 & \bluedelta{+0.4}
                 & \bluedelta{+0.9}
                 & \bluedelta{+0.6}
                 & \bluedelta{+0.2}
                 & \bluedelta{+0.6}
                 & \bluedelta{+18} \\
\midrule
\rowcolor{blue!10}
+ RL-Stage3     & 61.8 & 69.8 & 29.3 & 42.1 & 84.2 & 89.3 & 53.2 & 64.4 & 2376 \\
$\Delta$         & \bluedelta{+0.7} 
                 & \bluedelta{+1.0}
                 & \bluedelta{+1.7}
                 & \bluedelta{+0.4}
                 & \bluedelta{+0.4}
                 & \bluedelta{+0.3}
                 & \bluedelta{+0.3}
                 & \bluedelta{+0.5}
                 & \bluedelta{+8} \\
\bottomrule
\end{tabular}
}
\caption{Perfomance of our O$^2$-CritiCuRL across each training stage.}
\label{tab:stages-results}
\vspace{-10pt}
\end{table*}

\subsubsection{Hyperparameter $\alpha, \beta, \gamma$ Sensitivity Analysis}
We conduct an ablation study on the three weighting coefficients, $\alpha$, $\beta$, and $\gamma$, in the reward function used during the online update stage. As shown in Table \ref{tab:hyperparameter_grid}, different combinations of $\alpha$, $\beta$, and $\gamma$ have only a marginal impact on the model performance across various benchmarks. The performance variation is consistently within 0.6, with average fluctuations of only 0.2. These results indicate that our method is insensitive to reward weights, highlighting that its effectiveness mainly derives from the structural design rather than hyperparameter tuning.
\begin{table}[!htbp]
\vspace{-5pt}
\centering
\setlength{\tabcolsep}{0.2pt}
\renewcommand{\arraystretch}{1.2}
\resizebox{\columnwidth}{!}{
\fontsize{12pt}{14pt}\selectfont
\begin{tabular}{cccccccc}
\toprule
\multicolumn{3}{c}{\textbf{Parameter}} 
& \multirow{2}{*}{\textbf{M-Vista\,}} 
& \multirow{2}{*}{\textbf{M-Vision\,}} 
& \multirow{2}{*}{\textbf{SciQA\,}} 
& \multirow{2}{*}{\textbf{MMMU\,}} 
& \multirow{2}{*}{\textbf{MMStar}} \\
\cmidrule(lr){1-3}
\textbf{$\alpha$} &\textbf{$\beta$} &\textbf{$\gamma$} &  &  &  &  &  \\
\midrule
1 & 1 & 1 & 69.6 & 29.4 & 89.1 & 53.0 & 64.0 \\
2 & 1 & 1 & 69.8 & 29.3 & 89.3 & 53.2 & 64.4 \\
1 & 2 & 1 & 69.9 & 28.9 & 89.2 & 52.8 & 64.2 \\
2 & 2 & 1 & 69.5 & 29.2 & 89.0 & 53.0 & 63.8 \\
1 & 1 & 1 & 69.3 & 29.3 & 89.4 & 52.9 & 64.0 \\
2 & 1 & 2 & 69.8 & 29.2 & 89.2 & 53.1 & 64.2 \\
\bottomrule
\end{tabular}
}
\caption{Hyperparameter sensitivity analysis.}
\label{tab:hyperparameter_grid}
\vspace{-20pt}
\end{table}
\vspace{-10pt}

\subsubsection{Critical Step Number $k$ Selection}
\vspace{-5pt}
As mentioned in Sec \ref{sec:offline-step-distillation}, for efficiency considerations, we adopt an adaptive $k$ rather than a fixed $k$ for selecting the number of critical steps. To validate this design, we compare $adaptive-k$ with several fixed-$k$ settings on Math, Science, and General benchmarks. We use $\frac{\Delta\text{Avg}}{\text{GPU Hours}}$, where $\Delta\text{Avg}$ denotes the accuracy gain from cold-start SFT to the final RL model, to measure the trade-off between performance improvement and RL training cost. As shown in Table~\ref{tab:step_number_k}, $k$=1 yields limited gains, while fixed $k$=2 or $k$=3 brings only marginal improvement but substantially increases computation. In contrast, adaptive-$k$ achieves the best accuracy-cost trade-off.
\begin{table}[htbp]
\vspace{-8pt}
\centering
\setlength{\tabcolsep}{0.2pt}
\renewcommand{\arraystretch}{1.2}
\resizebox{\columnwidth}{!}{
\fontsize{12pt}{14pt}\selectfont
\begin{tabular}{cccccc}
\toprule
\textbf{$k$} & \textbf{M-Vista\,} & \textbf{M-Vision\,} & \textbf{SciQA\,} & \textbf{MMStar\,} &\textbf{$\tfrac{\Delta\text{Avg} \,{\scriptstyle(\times 10^{-3})}}{\text{GPU Hours}}\uparrow$}\\
\midrule
$k$=1 & 68.1 & 28.9 & 88.4 & 64.0 & 14.75 \\
$k$=2 & 69.7 & 29.2 & 89.0 & 64.4 &  11.60\\
$k$=3 & 69.6 & \textbf{29.4} & 89.2 & \textbf{64.5} & 8.01\\
\textbf{Adaptive-$k$}& \textbf{69.8} & 29.3 & \textbf{89.3} & 64.4 & \textbf{16.33}\\
\bottomrule
\end{tabular}
}
\caption{Critical Step Number $k$ Selection.}
\label{tab:step_number_k}
\vspace{-30pt}
\end{table}
\vspace{-10pt}

\subsubsection{Offline Rollout Number $n$ Selection}
As discussed in Sec \ref{sec:offline-step-distillation}, each sample is rolled out $n$ times in the offline stage for critical-step identification, and we set $n$=16 by default. We further ablate $n$ and use $\frac{\Delta\text{Avg}}{\text{GPU Hours}}$ to evaluate efficiency. As shown in Table~\ref{tab:rollout_number_n}, smaller values of $n$ lead to weaker gains, likely due to higher randomness and biased critical-step selection. Increasing $n$ to 32 brings only marginal improvement but substantially increases rollout cost. These results indicate that $n$=16 offers the best trade-off between accuracy and efficiency.
\begin{table}[htbp]
\vspace{-5pt}
\centering
\setlength{\tabcolsep}{0.2pt}
\renewcommand{\arraystretch}{1.2}
\resizebox{\columnwidth}{!}{
\fontsize{12pt}{14pt}\selectfont
\begin{tabular}{cccccc}
\toprule
\textbf{$n$} & \textbf{M-Vista\,} & \textbf{M-Vision\,} & \textbf{SciQA\,} & \textbf{MMStar\,} &\textbf{$\tfrac{\Delta\text{Avg} \,{\scriptstyle(\times 10^{-3})}}{\text{GPU Hours}}\uparrow$}\\
\midrule
4 & 67.8 & 28.2 & 88.8 & 63.6 & 12.44 \\
8 & 68.1 & 28.5 & 89.1 & 63.4 & 12.73 \\
\textbf{16}& \textbf{69.8} & \textbf{29.3} & 89.3 & 64.4 & \textbf{16.33} \\
32 & 69.7 & 29.2 & \textbf{89.5} & \textbf{64.5} & 13.70 \\
\bottomrule
\end{tabular}
}
\caption{Offline Rollout Number $n$ Selection.}
\label{tab:rollout_number_n}
\vspace{-20pt}
\end{table}
\vspace{-15pt}

\subsubsection{Critical-Step Metrics Ablation}

As shown in Table \ref{tab:critical-step-ablation}, we compare three candidate metrics for critical-step identification: the direct probability difference 
$\Delta p$, the normalized probability increment $\frac{p_i - p_{i-1}}{1 - p_{i-1}}$, and our proposed KL-based metric $K_i$. The results show that $K_i$ achieves the best performance on all benchmarks. These results suggest that the KL-based metric captures the overall shift of the output distribution after introducing a given step, and is therefore better suited to identifying steps that genuinely alter the reasoning trajectory and materially affect final correctness. The advantage of $K_i$ is particularly evident on tasks such as MathVista and MathVision, where successful prediction relies more heavily on coherent intermediate reasoning. Overall, this ablation study verifies the effectiveness of our critical-step metric for critical-step mining and provides direct support for the design of our offline critical-step selection strategy.
\begin{table}[htbp]
\vspace{-10pt}
\centering
\setlength{\tabcolsep}{0.2pt}
\renewcommand{\arraystretch}{1.2}
\resizebox{\columnwidth}{!}{
\fontsize{12pt}{14pt}\selectfont
\begin{tabular}{lccccc}
\toprule
\textbf{Metric\,} 
& \textbf{M-Vista\,} 
& \textbf{M-Vision\,} 
& \textbf{ChartQA\,} 
& \textbf{SciQA\,} 
& \textbf{MMStar} \\
\midrule
$\Delta p$               
& 68.3 & 28.2 & 83.5 & 88.2 & 63.9 \\
$\frac{p_i - p_{i-1}}{1 - p_{i-1}}$  
& 68.9 & 28.5 & 83.7 & 88.9 & 64.2 \\
\textbf{$K_i$}                  
& \textbf{69.8} & \textbf{29.3} & \textbf{84.2} 
& \textbf{89.3} & \textbf{64.4} \\
\bottomrule
\end{tabular}
}
\caption{Ablations on Critical-Step Metrics.}
\label{tab:critical-step-ablation}
\vspace{-20pt}
\end{table}
\vspace{-20pt}

\subsubsection{Step-Acc Reward Ablation}
As shown in Table~\ref{tab:step-reward-ablation}, we evaluate the effect of the step-accuracy reward. Removing StepAcc leads to lower performance on all benchmarks, while incorporating it improves the results to 69.8, 29.3, 84.2, 89.3, and 64.4 on MathVista, MathVision, ChartQA, ScienceQA, and MMStar, with an average gain of 1.14. These consistent improvements show that supervising critical intermediate steps provides benefits beyond answer-level rewards. This is important because a correct final answer does not necessarily imply a valid reasoning process, and answer-only supervision may still encourage shortcut-based solutions. By explicitly rewarding critical-step correctness, StepAcc helps constrain intermediate reasoning and improves the robustness of end-task performance.
\begin{table}[htbp]
\vspace{-10pt}
\centering
\setlength{\tabcolsep}{0.2pt}
\renewcommand{\arraystretch}{1.2}
\resizebox{\columnwidth}{!}{
\fontsize{12pt}{14pt}\selectfont
\begin{tabular}{lccccc}
\toprule
\textbf{Setting\,} 
& \textbf{M-Vista\,} 
& \textbf{M-Vision\,}
& \textbf{ChartQA\,}
& \textbf{SciQA\,} 
& \textbf{MMStar} \\
\midrule
w/o StepAcc            
& 67.5 & 28.6 & 83.3 & 88.1 & 63.7 \\
\textbf{w/ StepAcc}       
& \textbf{69.8} & \textbf{29.3} & \textbf{84.2} 
& \textbf{89.3} & \textbf{64.4} \\
\bottomrule
\end{tabular}
}
\caption{Ablations on Step-Acc Reward.}
\label{tab:step-reward-ablation}
\vspace{-20pt}
\end{table}
\vspace{-20pt}

\subsubsection{Necessity of the Offline-Online Iteration}
We further assess the necessity of the offline-online iteration by comparing four settings: \emph{Fixed Key Steps}, where critical steps are identified once for one-time distillation; \emph{Only Offline Distillation}, which uses distilled supervision for SFT; \emph{Only Online RL}, i.e., vanilla GRPO without offline critical-step mining; and our full \emph{Offline--Online Iteration}. As shown in Table \ref{tab:offline_online_iteration}, the full method achieves the best performance across all benchmarks. The inferior results of \emph{Only Offline Distillation} indicate that static distilled supervision alone cannot fully translate critical-step information into strong reasoning ability. Similarly, \emph{Only Online RL} underperforms the full method, showing that RL without offline critical-step guidance relies on weaker and less structured rewards. Moreover, although \emph{Fixed Key Steps} improves over the single-stage baselines, it remains consistently worse than our iterative design, suggesting that critical steps should be updated as the model evolves. These results demonstrate that the effectiveness of our framework comes from the closed offline--online loop, where offline distillation refreshes critical-step supervision and online optimization improves the model for more accurate subsequent step discovery.

\begin{table}[htbp]
\vspace{-5pt}
\centering
\setlength{\tabcolsep}{0.2pt}
\renewcommand{\arraystretch}{1.2}
\resizebox{\columnwidth}{!}{%
\fontsize{12pt}{14pt}\selectfont
\begin{tabular}{lccccc}
\toprule
\textbf{Method\,} & \textbf{M-Vista\,} & \textbf{M-Vision\,} & \textbf{ChartQA\,} & \textbf{SciQA\,} & \textbf{MMStar} \\
\midrule
Fixed Key Steps & 68.1 & 28.9 & 83.8 & 88.4 & 64.0 \\
Only Offline Distillation & 64.0 & 23.7 & 82.3 & 87.9 & 62.5 \\
Only Online RL & 67.5 & 28.6 & 83.3 & 88.1 & 63.7 \\
\textbf{Offline-Online Iteration} & \textbf{69.8} & \textbf{29.3} & \textbf{84.2} & \textbf{89.3} & \textbf{64.4} \\
\bottomrule
\end{tabular}%
}
\caption{Necessity of the offline-online iteration.}
\label{tab:offline_online_iteration}
\vspace{-20pt}
\end{table}
\vspace{-10pt}

\begin{figure*}[!ht]
    \centering
    \includegraphics[width=1.0\linewidth]{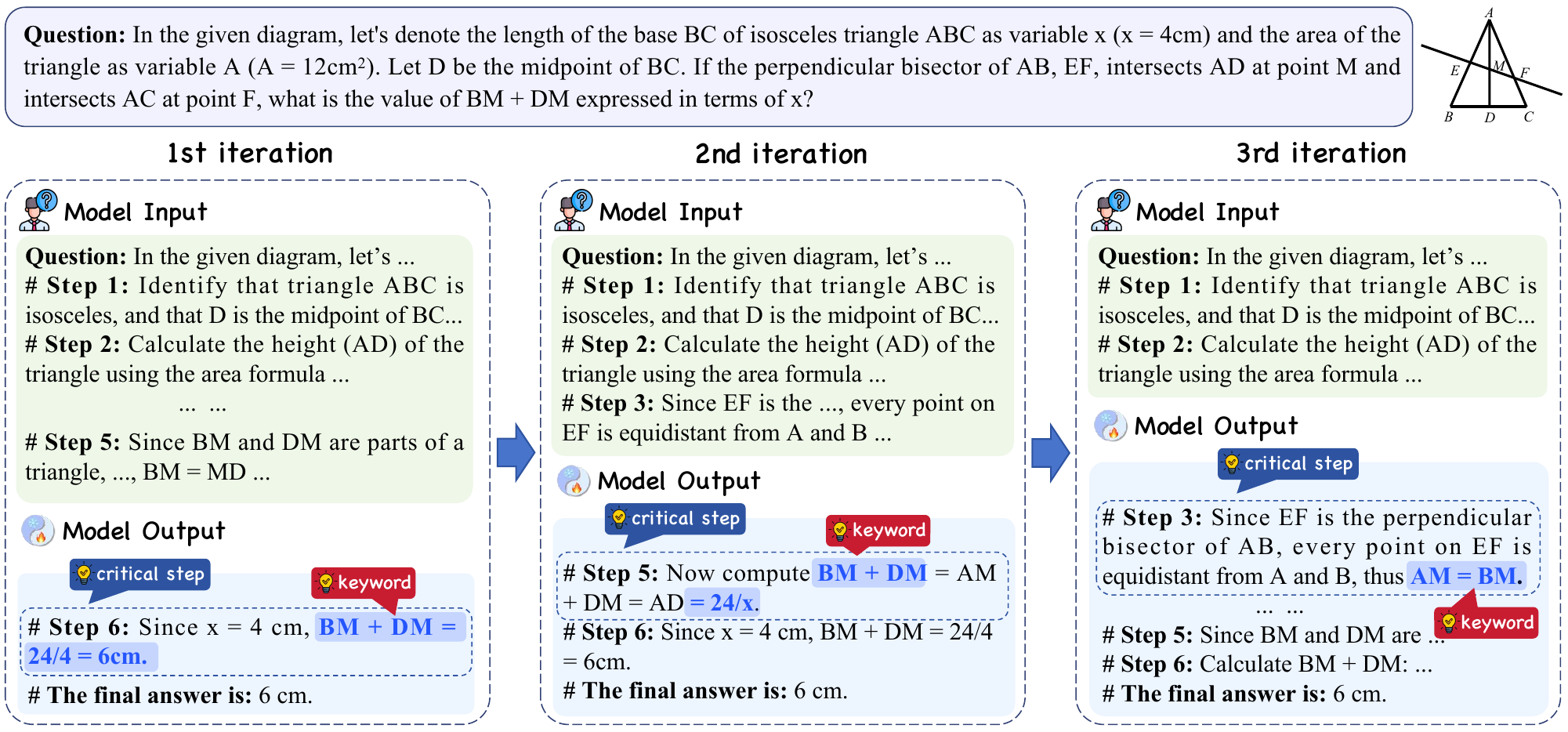}
    \caption{The critical step shifts across iterations.}
    \label{fig:critical_step_shifts}
\end{figure*}

\subsubsection{Offline Rollout Efficiency Analysis}
Since our framework introduces an additional offline rollout stage, it inevitably incurs extra computation compared with vanilla GRPO. To assess whether this added cost is justified, we conduct efficiency comparisons on MathVista, ChartQA, and ScienceQA. We use the total training FLOPs to quantify computational overhead, and adopt $\frac{\Delta \text{Avg}}{\text{FLOPs}}$ as the efficiency metric, where $\Delta \text{Avg}$ denotes the improvement of the final average performance over the cold-start model. As shown in Table \ref{tab:efficiency_analysis}, our method indeed introduces additional cost. In return, the performance gain is substantial: the average score on the three benchmarks improves from 79.0 to 81.1, and the efficiency metric $\frac{\Delta \text{Avg}}{\text{FLOPs}}$ increases markedly from 32.03 to 51.63. These results show that the added offline stage is highly cost-effective: forward-only rollouts provide stronger critical-step supervision with limited overhead, leading to a better computation \& performance trade-off.
\begin{table}[htbp]
\vspace{-5pt}
\centering
\setlength{\tabcolsep}{0.5pt}
\renewcommand{\arraystretch}{1.2}

\resizebox{\columnwidth}{!}{%
\fontsize{12pt}{14pt}\selectfont
\begin{tabular}{lccccccc}
\toprule
\multirow{2}{*}{\textbf{Method}\,}
& \multirow{2}{*}{\textbf{M-Vista}\,}
& \multirow{2}{*}{\textbf{ChartQA}\,}
& \multirow{2}{*}{\textbf{SciQA}\,}
& \multirow{2}{*}{\textbf{Avg}$\uparrow$}
& \multicolumn{2}{c}{
        \textbf{FLOPs} $\downarrow$
        {\scriptsize $(\times 10^{18})$}
}
& \multirow{2}{*}{\textbf{$\tfrac{\Delta\text{Avg}\,{\scriptstyle(\times10^{-3})}}{\text{FLOPs}} \uparrow$}} \\
\cmidrule(lr){6-7}
& & & & &
\textbf{Offline}
& \textbf{Online}
& \\
\midrule
GRPO
& 67.5
& 83.3
& 88.1
& 79.1
& --
& 43.7187
& 32.03 \\

\textbf{Ours}
& \textbf{69.8}
& \textbf{84.2}
& \textbf{89.3}
& \textbf{81.1}
& 6.0396
& 48.1998
& \textbf{51.63} \\
\bottomrule
\end{tabular}%
}

\caption{Offline rollout efficiency analysis.}
\label{tab:efficiency_analysis}
\vspace{-20pt}
\end{table}
\vspace{-20pt}

\subsection{Discussion}\label{sec:discussion}

We offer the following insights based on our observations.

\noindent
\textbf{Critical steps are more important than long CoT}: As shown in Table \ref{tab:vlm-results}.
R1-VL incorporates step-level rewards, but its performance remains far below ours ($6.6\%\downarrow$), showing that naively rewarding all steps is ineffective.
This contrast highlights that reasoning quality is determined not by longer chains of thought, but by accurately identifying, distilling, and reinforcing the few critical steps that truly matter.\\
\textbf{Critical steps are dynamic, not fixed}: As shown in Figure~\ref{fig:critical_step_shifts}, the steps regarded as critical change during training. 
When the model’s reasoning ability improves, previously difficult steps may no longer limit performance, while new key steps become more important. 
This observation suggests that fixed ``gold'' reasoning traces are not universally optimal. 
Instead, step-level supervision should evolve with the model’s capability~\citep{qiu2025disentangled}.
Across multiple benchmarks, early training tends to select steps that are close to the final answer, because they provide direct reward improvement. 
As the model becomes stronger, the selected steps gradually move toward intermediate conclusions or important theorems that support the derivation. 
Redundant or repetitive steps are rarely selected. 
This pattern shows that our framework is able to identify the model’s current reasoning bottlenecks and assign supervision to the steps that are most beneficial for further optimization.\\
\textbf{O$^2$-CritiCuRL yields more stable RL}: As shown in Figure~\ref{fig:training_curves}, raw step-wise and answer-level rewards fluctuate substantially, while their smoothed curves increase steadily. This indicates that offline–online separation reduces reward variance and improves credit assignment.

\section{Case Study}

\begin{figure*}[!htbp]
    \centering
    \includegraphics[width=\textwidth]{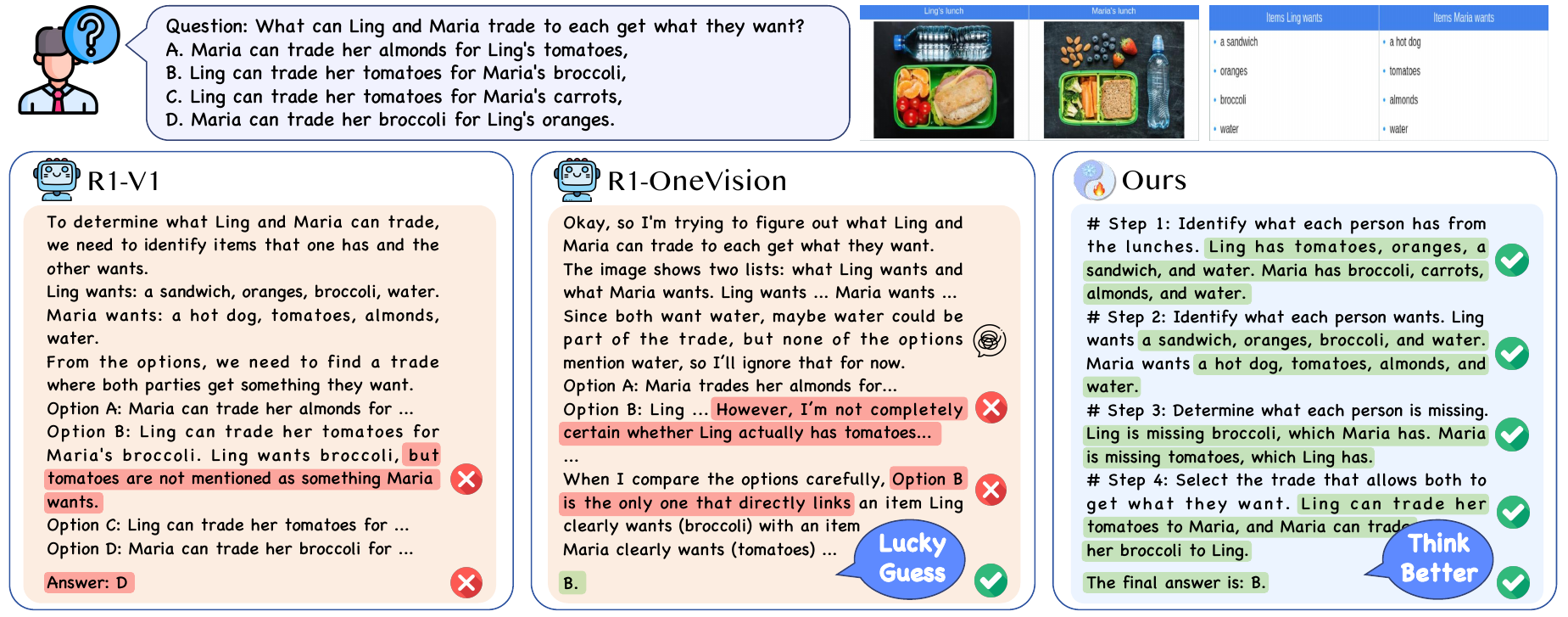}
    \caption{\textbf{Case Study on ScienceQA}: The reasoning behaviors of Vision-R1, R1-VL, and our proposed O²-CritiCuRL. Existing Vision-R1 provides incorrect answers due to flawed intermediate reasoning, while R1-VL reaches the correct answer only through a ``lucky guess''. In contrast, our O²-CritiCuRL identifies and follows the correct critical steps, yielding logically sound and interpretable output.}
    \label{fig:case_study_scienceqa}
\end{figure*}

Figure~\ref{fig:case_study_scienceqa} presents a ScienceQA example that requires matching the items owned and desired by two individuals. R1-V1 overlooks the fact that Maria wants tomatoes and consequently selects an incorrect trade. R1-OneVision reaches the correct answer, but its reasoning remains uncertain about whether Ling has tomatoes, indicating that the prediction is largely a lucky guess. In contrast, O$^2$-CritiCuRL systematically identifies what each person has and wants, determines that Ling needs broccoli while Maria needs tomatoes, and selects the mutually beneficial trade. This example shows that our method reaches the correct answer through a complete and logically grounded reasoning process.

\begin{figure*}[!ht]
    \centering
    \includegraphics[width=\textwidth]{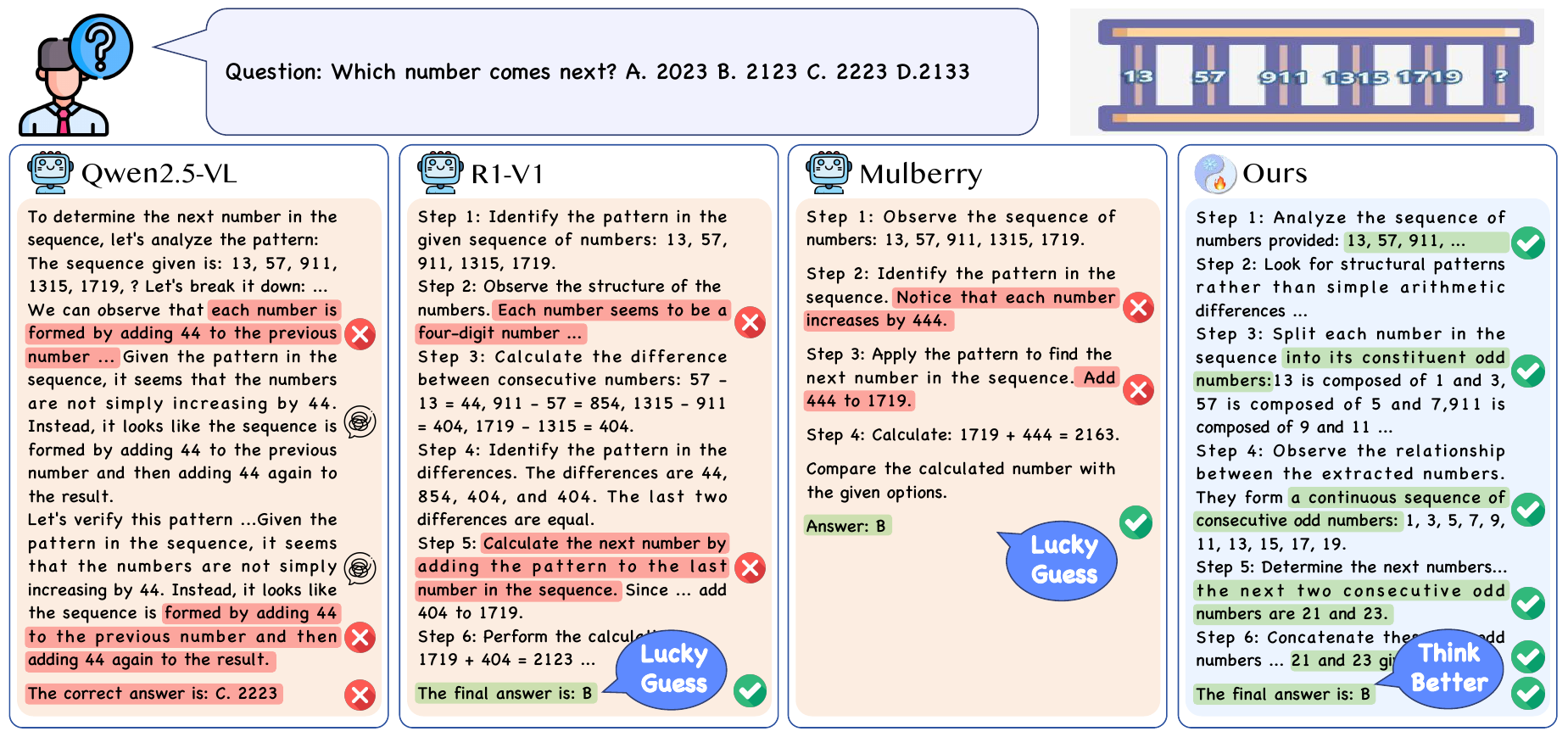}
    \caption{\textbf{Case Study on MMStar}: The reasoning behaviors of Vision-R1, R1-VL, and our proposed O²-CritiCuRL. Existing Vision-R1 provides incorrect answers due to flawed intermediate reasoning, while R1-VL reaches the correct answer only through a ``lucky guess''. In contrast, our O²-CritiCuRL identifies and follows the correct critical steps, yielding logically sound and interpretable output.}
    \label{fig:case_study_mmstar}
    \vspace{-20pt}
\end{figure*}

Figure~\ref{fig:case_study_mmstar} shows an MMStar number-sequence problem in which the numbers are formed by concatenating consecutive odd integers. Qwen2.5-VL incorrectly interprets the sequence as being generated through repeated arithmetic increments and therefore produces the wrong answer. Although R1-V1 and Mulberry select the correct option, their intermediate calculations rely on inconsistent difference patterns and do not recover the underlying structure of the sequence. By contrast, O$^2$-CritiCuRL decomposes each term into its constituent odd numbers, identifies the continuous sequence $1,3,5,\ldots,19$, and correctly infers that the next term is formed by concatenating $21$ and $23$. The resulting answer is therefore supported by the actual generative rule rather than an accidental numerical match.

\begin{figure*}[!htbp]
    \centering
    \includegraphics[width=\textwidth]{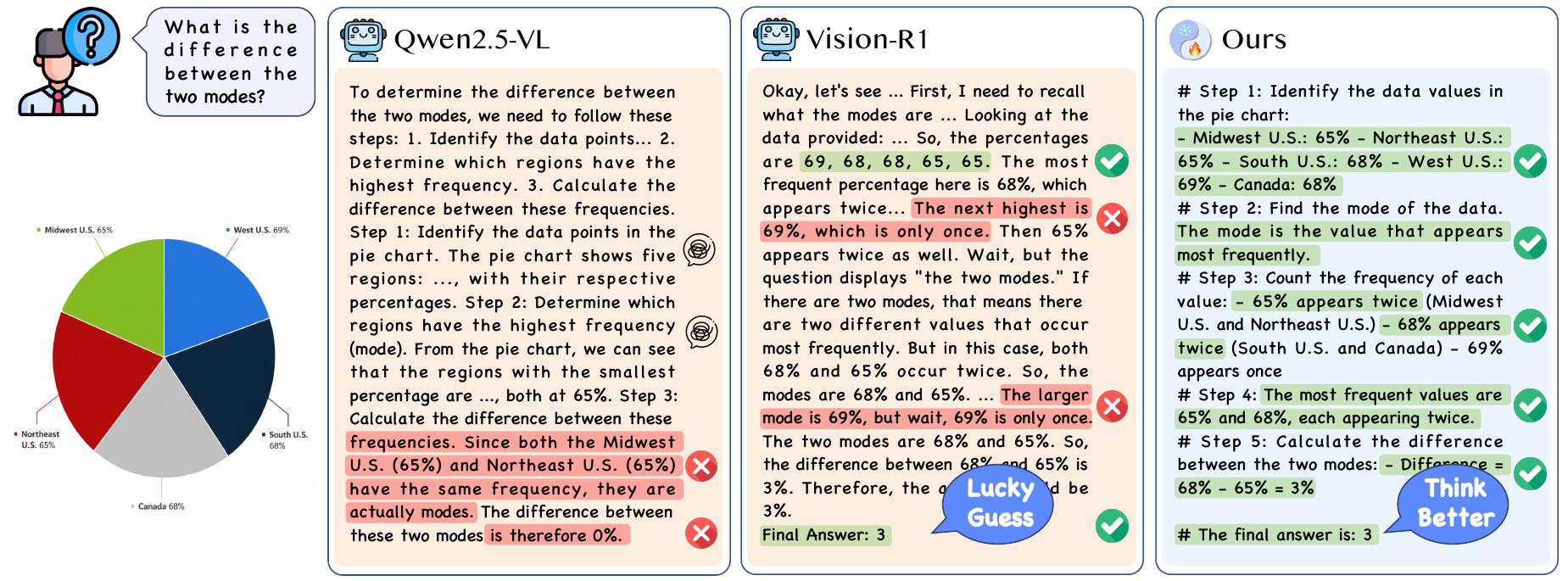}
    \caption{\textbf{Case Study on ChartQA}: The reasoning behaviors of Vision-R1, R1-VL, and our proposed O²-CritiCuRL. Existing Vision-R1 provides incorrect answers due to flawed intermediate reasoning, while R1-VL reaches the correct answer only through a ``lucky guess''. In contrast, our O²-CritiCuRL identifies and follows the correct critical steps, yielding logically sound and interpretable output.}
    \vspace{-15pt}
    \label{fig:case_study_chartqa}
\end{figure*}

Figure~\ref{fig:case_study_chartqa} presents a ChartQA example that asks for the difference between the two modes of the values shown in a pie chart. Qwen2.5-VL incorrectly treats the two occurrences of $65\%$ as two separate modes and concludes that their difference is zero. Vision-R1 eventually gives the correct answer, but its reasoning contains contradictory statements about which values constitute the modes, again suggesting a lucky guess. O$^2$-CritiCuRL first extracts all values from the chart, counts their frequencies, correctly identifies $65\%$ and $68\%$ as the two modes, and computes their difference as $3\%$. This case demonstrates that our method maintains consistency between visual evidence, intermediate reasoning, and the final answer.

Overall, these case studies demonstrate a consistent distinction between answer correctness and reasoning correctness. Baseline models either produce incorrect answers because of flawed intermediate steps or occasionally reach correct answers through inconsistent or unsupported reasoning. In contrast, O$^2$-CritiCuRL identifies and executes the decisive reasoning steps in each task, producing solutions that are not only accurate but also concise, logically coherent, and interpretable. Rather than merely matching the final answer, our method preserves consistency between the observed evidence, intermediate deductions, and the resulting prediction. These findings further suggest that critical-step-aware supervision can improve both reasoning faithfulness and robustness across diverse multimodal settings. Additional case studies are provided in the appendix to further illustrate this behavior.

\section{Conclusion}
In this work, we propose O²-CritiCuRL, an offline–online curriculum reinforcement learning framework that emphasizes critical-step awareness. By adaptively identifying decisive steps and iteratively refining them through offline decomposition and online reinforcement learning, our method improves accuracy, interpretability, and efficiency over conventional step-level RL. We expect that the principle of critical-step awareness is not limited to mathematical reasoning, but can be broadly applied to domains that demand transparent and trustworthy decision-making, such as scientific discovery, legal analysis, and medical diagnosis. Future work may integrate human feedback into step identification on larger multimodal datasets, and design adaptive curricula that co-evolve in real-world cases under practical tasks.

\bibliographystyle{plainnat}
\bibliography{main}

\begin{appendices}

\section{Ethics Statement}
In this study, no human subjects or animal experimentation was involved. 
All datasets used were sourced in compliance with relevant usage guidelines, ensuring no violation of privacy. 
We have taken care to avoid any biases or discriminatory outcomes in our research process. 
No personally identifiable information was used, and no experiments were conducted that could raise privacy or security concerns. 
We are committed to maintaining transparency and integrity throughout the research process.

\section{Reproducibility Statement}
We have made every effort to ensure that the results presented in this paper are reproducible. 
The experimental setup, including training steps, model configurations, and hardware details, is described in detail in the paper, 
We also provide a full description of experiments, to assist others in reproducing our experiments.
Additionally, datasets, such as MathVista, MathVision are publicly available, ensuring consistent and reproducible evaluation results.
We believe these measures will enable other researchers to reproduce our work and further advance the field.

\section{Theoretical Foundation and Interpretation}
This section provides the theoretical foundation, detailing the derivation and practical implementation of the criticality metric $\mathcal{K}_i$.
In our approach, the criticality of a reasoning step is assessed based on the change in the model's accuracy after sequentially introducing each step. Assuming each step is correct, adding a step will not decrease the model's accuracy; it can either improve it or leave it unchanged (if the step provides no additional information).

Introducing a critical step inevitably causes a significant shift in the answer probability distribution: the probability of the correct answer increases, while the probabilities of other answers decrease. To describe this process in a simple yet principled way, we introduce the \textbf{uniform compression assumption}:
\begin{enumerate}
    \item When the model has no reasoning ability, all answers are assigned approximately equal initial probabilities; if the answer space is very large, these probabilities naturally approach zero.
    \item When a critical step is introduced, the correct answer probability increases, while the remaining probability mass of other answers is assumed, in an idealized sense, to be uniformly redistributed. This assumption provides a tractable and reasonable approximation of the probability shift.
\end{enumerate}

To quantify the difference between the prior and posterior distributions induced by a reasoning step, we use the \textbf{Kullback-Leibler (KL) divergence}, defined as
\begin{equation}
D_{\text{KL}}(P \parallel Q) = \sum_{\omega \in \Omega} P(\omega) \log \frac{P(\omega)}{Q(\omega)}.
\end{equation}
Its fundamental interpretation is the additional information required to represent the target distribution $P$ using a code optimized for the distribution $Q$. In our context, the KL divergence measures the extra information needed for the current distribution to align with a distribution that successfully leads to the correct answer.

Within this framework, the criticality of step $S_i$ is quantified as the KL divergence between the posterior $P_i$ and the prior $P_{i-1}$:
\begin{equation}
\mathcal{K}_i = D_{\text{KL}}(P_i \parallel P_{i-1}).
\end{equation}
Intuitively:
\begin{itemize}
    \item \textbf{Large $\mathcal{K}_i$}: Indicates a substantial difference between $P_i$ and $P_{i-1}$, meaning the step $S_i$ provides significant new information, which is highly critical for correcting the model when it was previously far from the correct answer.
    \item \textbf{Small $\mathcal{K}_i$}: Indicates that $P_i$ is similar to $P_{i-1}$, meaning the step $S_i$ offers only minor refinements, acting more like a fine-tuning adjustment when the model is already close to the correct answer.
\end{itemize}

\section{Derivation of the Computable Metric}

The definition of $\mathcal{K}_i$ is general but computationally complex as it requires a sum over the entire answer space $\Omega$. To derive a tractable metric, we introduce a structural assumption about how reasoning steps affect the probability distribution.

We begin by decomposing the sum into the contribution from the correct answer $\omega^*$ and the space of all incorrect answers:
\begin{equation}
\mathcal{K}_i = P_i(\omega^*)\log\frac{P_i(\omega^*)}{P_{i-1}(\omega^*)} + \sum_{\omega \neq \omega^*} P_i(\omega)\log\frac{P_i(\omega)}{P_{i-1}(\omega)}.
\end{equation}

Let $p_i = P_i(\omega^*)$ denote the probability assigned to the correct answer at step $i$. Consequently, the total probability mass of all incorrect answers is $1 - p_i = \sum_{\omega \neq \omega^*} P_i(\omega)$.

\textbf{Coarse-Grained Approximation and Residual Term.}
Introducing a critical step induces a shift in the model's answer distribution: the probability mass assigned to the correct answer increases, while the probability mass over incorrect answers is adjusted accordingly. To quantify the distributional change caused by step $S_i$, we use the Kullback-Leibler (KL) divergence:
$
D_{\mathrm{KL}}(P|Q)=\sum_{\omega\in\Omega}P(\omega)\log\frac{P(\omega)}{Q(\omega)}.
$
Within our framework, the criticality of step $S_i$ is measured by the distributional shift between the posterior distribution $P_i$ and the prior distribution $P_{i-1}$:
$
K_i=D_{\mathrm{KL}}(P_i|P_{i-1})=
\sum_{\omega\in\Omega}
P_i(\omega)\log\frac{P_i(\omega)}{P_{i-1}(\omega)}.
$

Let $\omega^*$ denote the correct answer, and let $p_i=P_i(\omega^*)$ be the probability assigned to the correct answer after introducing step $S_i$. We first decompose the KL divergence into the contribution from the correct answer and the contribution from the incorrect-answer subspace:
$
K_i
=
p_i\log\frac{p_i}{p_{i-1}}
+
\sum_{\omega\neq\omega^*}
P_i(\omega)\log\frac{P_i(\omega)}{P_{i-1}(\omega)}.
$
To characterize step-level distribution changes in multimodal reasoning, we further decompose the answer-space shift into a coarse-grained correctness term and a residual term within the incorrect-answer subspace. The former measures how much probability mass is shifted toward the correct answer, while the latter captures the redistribution among incorrect answers, including possible semantic clusters caused by visual perception errors, reasoning shortcuts, or similar answer candidates.

Specifically, we define the normalized distribution over the incorrect-answer subspace as
$
Q_i(\omega)=\frac{P_i(\omega)}{1-p_i},
\quad \omega\neq\omega^* .
$
Thus, for each incorrect answer $\omega\neq\omega^*$, we have
$
P_i(\omega)=(1-p_i)Q_i(\omega).
$
Substituting this decomposition into the incorrect-answer term gives
\begin{equation}
\begin{aligned}
&\sum_{\omega\neq\omega^*}
P_i(\omega)\log\frac{P_i(\omega)}{P_{i-1}(\omega)} \\
&=
\sum_{\omega\neq\omega^*}
(1-p_i)Q_i(\omega)
\log
\frac{(1-p_i)Q_i(\omega)}
{(1-p_{i-1})Q_{i-1}(\omega)}\\
&=
(1-p_i)\log\frac{1-p_i}{1-p_{i-1}}
+
(1-p_i)D_{\mathrm{KL}}(Q_i|Q_{i-1}).
\end{aligned}
\end{equation}
Therefore, the full answer-space KL divergence can be written as
\begin{equation}
    \begin{aligned}
K_i
&=
p_i\log\frac{p_i}{p_{i-1}}
+
(1-p_i)\log\frac{1-p_i}{1-p_{i-1}} \\
&\quad+
(1-p_i)D_{\mathrm{KL}}(Q_i|Q_{i-1}).
    \end{aligned}
\end{equation}
The last term explicitly represents the internal redistribution among incorrect answers. Therefore, the formulation does not require incorrect answers to be uniformly distributed; instead, semantic clustering among wrong candidates is naturally absorbed into the residual term.

In practice, our offline rollout estimates the correctness-level transition after each reasoning prefix. Accordingly, we use the coarse-grained correctness term as the practical criticality metric:
\begin{equation}
K_i^{\mathrm{cg}}=
p_i\log\frac{p_i}{p_{i-1}}
+
(1-p_i)\log\frac{1-p_i}{1-p_{i-1}}.
\end{equation}
This term is exactly the KL divergence between two Bernoulli distributions over correctness, i.e., whether the model reaches the correct answer or not. Since
$
D_{\mathrm{KL}}(Q_i|Q_{i-1})\geq 0,
$
we have
$
K_i^{\mathrm{cg}}
\leq
D_{\mathrm{KL}}(P_i|P_{i-1}).
$
Thus, the coarse-grained metric can be interpreted as a conservative approximation of the full answer-space distribution shift. It focuses on the probability mass transferred toward the correct answer, while preserving the effect of incorrect-answer clustering as a residual component.

Combining these contributions yields our final computable metric for step criticality:
\begin{equation}
\boxed{
\mathcal{K}_i = p_i \log \frac{p_i}{p_{i-1}} + (1 - p_i) \log \frac{1 - p_i}{1 - p_{i-1}}
}.
\end{equation}

\section{Boundary Behavior, Robustness, and Smoothing}

The metric $\mathcal{K}_i$ exhibits extreme and theoretically justified behavior at the boundaries of the probability space, which necessitates careful handling in practice.

\textbf{Boundary Analysis:}

\paragraph{Case 1: $p_{i-1} \to 0^+$}  
Consider
\begin{equation}
\mathcal{K}_i = p_i \log \frac{p_i}{p_{i-1}} + (1 - p_i) \log \frac{1 - p_i}{1 - p_{i-1}}.
\end{equation}

As $p_{i-1} \to 0^+$, the second term behaves as
\begin{equation}
\begin{aligned}
(1 - p_i)\log\frac{1 - p_i}{1 - p_{i-1}}
&= (1 - p_i)\log\!\left(1 + \frac{p_{i-1} - p_i}{1 - p_{i-1}}\right) \\
&\to (1 - p_i)\log(1 - p_i).
\end{aligned}
\end{equation}

which remains finite. The first term behaves as
\begin{equation}
    p_i \log \frac{p_i}{p_{i-1}} = p_i (\log p_i - \log p_{i-1}) \sim - p_i \log p_{i-1} \to +\infty,
\end{equation}
since $-\log p_{i-1} \to +\infty$. Therefore, $\mathcal{K}_i \to +\infty$, reflecting an infinite information gain when updating from near-zero prior probability to a finite posterior.

\paragraph{Case 2: $p_{i-1} \to 1^-$}  
Let $q_i = 1 - p_i$ and $q_{i-1} = 1 - p_{i-1}$. Then
\begin{equation}
    \mathcal{K}_i = (1 - q_i) \log \frac{1 - q_i}{1 - q_{i-1}} + q_i \log \frac{q_i}{q_{i-1}}.    
\end{equation}
As $q_{i-1} \to 0^+$, the first term behaves as
\[
(1 - q_i) \log \frac{1 - q_i}{1 - q_{i-1}} \to \log 1 = 0,
\]  
and the second term
\begin{equation}
    q_i \log \frac{q_i}{q_{i-1}} = q_i (\log q_i - \log q_{i-1}) \sim - q_i \log q_{i-1} \to 0,    
\end{equation}
because $q_i \to 0$ along with $q_{i-1}$. Hence $\mathcal{K}_i \to 0$, indicating negligible information gain when the model is already nearly certain.

\textbf{Robustness Property via Taylor Expansion:}

Let $\Delta p = p_i - p_{i-1}$. Expanding $\mathcal{K}_i$ around $p_i = p_{i-1}$ using $\log(1+x) \approx x - \frac{x^2}{2} + \mathcal{O}(x^3)$:


\begin{equation}
\resizebox{0.99\linewidth}{!}{$
\begin{aligned}
\mathcal{K}_i 
&= p_{i-1} \log \frac{p_{i-1} + \Delta p}{p_{i-1}} 
  + (1 - p_{i-1}) \log \frac{1 - p_{i-1} - \Delta p}{1 - p_{i-1}} \\
&= p_{i-1} \log \left( 1 + \frac{\Delta p}{p_{i-1}} \right) 
  + (1 - p_{i-1}) \log \left( 1 - \frac{\Delta p}{1 - p_{i-1}} \right) \\
&\approx p_{i-1} \left( \frac{\Delta p}{p_{i-1}} - \frac{(\Delta p)^2}{2 p_{i-1}^2} \right) 
  + (1 - p_{i-1}) \left( - \frac{\Delta p}{1 - p_{i-1}} - \frac{(\Delta p)^2}{2 (1 - p_{i-1})^2} \right) \\
&= \Delta p - \frac{(\Delta p)^2}{2 p_{i-1}} - \Delta p - \frac{(\Delta p)^2}{2 (1 - p_{i-1})} \\
&= \frac{(\Delta p)^2}{2} \left( \frac{1}{1 - p_{i-1}} + \frac{1}{p_{i-1}} \right) \\
&= \frac{(\Delta p)^2}{2 p_{i-1}(1 - p_{i-1})}.
\end{aligned}
$}
\end{equation}

Summing both contributions:
\begin{equation}
    \mathcal{K}_i \approx \frac{(\Delta p)^2}{2} \left( \frac{1}{1 - p_{i-1}} + \frac{1}{p_{i-1}} \right) = \frac{(\Delta p)^2}{2 p_{i-1}(1 - p_{i-1})}.
\end{equation}

This shows that for small $\Delta p$, $\mathcal{K}_i$ is of order $(\Delta p)^2$, making it inherently robust to small fluctuations in the probability estimates.

\textbf{Smoothing Technique:}  
To ensure numerical stability, smoothing is applied at both ends of the probability range. At the lower end, if $p_i$ or $p_{i-1}$ equals 0, this is effectively a numerical error since the model's initial reasoning ability cannot be exactly zero. To avoid excessively large KL divergence values in practice, these probabilities are replaced by a small positive constant $\epsilon$, e.g., $\epsilon = 0.1$, which is much smaller than the minimal sampling probability precision.  

At the upper end, if $p_i$ or $p_{i-1}$ equals 1, although the result does not theoretically diverge, computing $\log 0$ could occur during calculations, which leads to numerical instability. Therefore, these probabilities are also capped below $1-\epsilon$ to ensure stability:  
\begin{equation}
\begin{aligned}
\tilde{p}_i &= \min\Big(\max(p_i, \epsilon), \, 1-\epsilon\Big), \\
\tilde{p}_{i-1} &= \min\Big(\max(p_{i-1}, \epsilon), \, 1-\epsilon\Big), \\
\epsilon &\ll \text{minimal sampling probability}.
\end{aligned}
\end{equation}

The smoothed criticality metric is then computed as:  
\begin{equation}
\mathcal{K}_i^{\text{(smooth)}} = \tilde{p}_i \log \frac{\tilde{p}_i}{\tilde{p}_{i-1}} + (1 - \tilde{p}_i) \log \frac{1 - \tilde{p}_i}{1 - \tilde{p}_{i-1}}.
\end{equation}

\section{Reward Function}

Our reward function is designed to guide the model towards three key objectives simultaneously:
\begin{enumerate}
    \item Correctly output the final answer.
    \item Perform step-wise, logical reasoning.
    \item Recognize whether the reasoning process is complete: if complete, directly output the answer; if incomplete, correctly continue the reasoning steps.
\end{enumerate}

The entire training process is divided into two distinct phases, each with its own reward function. They work in tandem to achieve our ultimate goal.

\paragraph{Offline Rollout Phase:}

This phase does not involve model training. Instead, it serves as an analytical tool to identify and evaluate critical steps within a Chain-of-Thought (CoT) demonstration. By using the reward function in this stage, we can prune redundant or less essential steps, creating a more concise and efficient training dataset.

\paragraph{Online Update Phase:}

This is the active model training phase. We use the refined dataset from the offline stage and provide real-time, granular feedback with the online reward function. This guides the model to learn and execute a correct reasoning process. This two-stage approach ensures that we first obtain high-quality training data and then use that data effectively to train the model.

\subsection{Offline Rollout Reward Function}

This phase is focused on evaluating the informational value of each step in a demonstration. The reward function here helps in identifying which steps are most crucial for reaching the correct final answer.

\paragraph{Answer Accuracy $R_{\text{acc}}$:}

This metric provides a foundational measure of the model's ability to produce the correct final answer after a given number of reasoning steps.
\begin{itemize}
    \item For each input sample, we execute multiple simulated "rollouts." In each rollout, we extract the final answer (the text following \texttt{\# The final answer is:}) and compare it to the ground truth.
    \begin{itemize}
        \item If the answer is correct, we assign a reward of $R_{\text{acc}} = 1$.
        \item If the answer is incorrect, the reward is $R_{\text{acc}} = 0$.
    \end{itemize}
    \item The mean correctness probability, $p_i$, is then calculated by averaging the $R_{\text{acc}}$ values over a series of $n$ rollouts. This is performed after step $i$ has been provided to the model, meaning the model has access to all steps from 1 to $i$.
\begin{equation}
    p_i = \frac{1}{n} \sum_{j=1}^{n} R_{\text{acc}}^{(j)}
\end{equation}
    Here, $p_i$ represents the mean probability that the model can correctly solve the problem after having access to the first $i$ reasoning steps.
\end{itemize}

\paragraph{Critical Step Metric $K_i$:}

The Critical Step Metric, $K_i$, is the core of our offline analysis. It quantifies the informational value of a specific step by measuring how significantly it alters the model's probability of producing the correct final answer.
\begin{itemize}
    \item We compute the Kullback-Leibler (KL) divergence between the correctness probabilities of consecutive steps, $p_i$ and $p_{i-1}$. A large $K_i$ value signifies that the step provides crucial new information that dramatically improves the chances of getting the final answer correct.
\begin{equation}
    K_i = p_i \log \frac{p_i}{p_{i-1}} + (1-p_i) \log \frac{1-p_i}{1-p_{i-1}}
\end{equation}
    Conversely, a small $K_i$ indicates the step is less essential or even redundant.
\end{itemize}

\subsection{Online Update Reward Function}

This phase involves granular, real-time feedback to guide the model towards producing a correct and well-formatted reasoning process.

\paragraph{Answer Accuracy Score $R_a$:}

This score evaluates whether the final answer produced by the model matches the ground-truth solution. As the most direct indicator of task success, it reflects the model's ability to arrive at the correct conclusion after completing its reasoning process.
\begin{itemize}
    \item If the model's final answer exactly matches the ground-truth answer (after normalization, such as trimming whitespace and removing irrelevant punctuation), then $R_a = 1$.
    \item If the final answer does not match the ground-truth answer, or if the model produces multiple conflicting answers, then $R_a = 0$.
    \item Numerical answers are compared using both symbolic and approximate matching to account for equivalent forms (e.g., $0.5$ vs.\ $\frac{1}{2}$), but any semantic mismatch results in $R_a = 0$.
\end{itemize}

\paragraph{Format Score $R_f$:}

This score ensures that the model's output strictly adheres to a predefined format, which is essential for consistent and automated evaluation throughout the training process.
\begin{itemize}
    \item If the model output strictly follows the specified format (\texttt{\# Step \# The final answer is: ...\textbackslash{}n}), $R_f = 1$.
    \item Any deviation from the required format results in $R_f = 0$.
\end{itemize}

\paragraph{Step Format $R_t$:}

This component evaluates whether the model's predicted step number aligns with the expected sequential order and whether it correctly decides to stop or continue the process. We use a variable, \texttt{judging\_step}, to denote the expected next step number that the model should output.
\begin{itemize}
    \item When reasoning is incomplete (\texttt{judging\_step $\neq$ 0}):
    \begin{itemize}
        \item Correct Continuation: If the model's output begins with \texttt{\# Step judging\_step}, it receives the highest reward, $R_t = 1$.
        \item Incorrect Step Number: If the model's output begins with \texttt{\# Step pred\_step} where \texttt{pred\_step} is not equal to \texttt{judging\_step}, the reward is calculated using a decay coefficient $k$: $R_t = \max(0, 1 - k \cdot |\text{pred\_step} - \text{judging\_step}|)$. The larger the deviation, the smaller the reward.
        \item Early Answer Output: If the model outputs the final answer before the reasoning is complete, the reward is $R_t = 0$.
    \end{itemize}
    \item When reasoning is complete (\texttt{judging\_step = 0}, indicating the model has reached the final step from the reference CoT):
    \begin{itemize}
        \item Correct Termination: If the model correctly outputs the final answer, it receives the highest reward, $R_t = 1$.
        \item Redundant Step: If the model generates a superfluous step after the reasoning is complete, it receives a small, token reward of $R_t = 0.05$. This encourages early answer output and reduces redundancy.
        \item Other Incorrect Output: Any other incorrect output results in $R_t = 0$.
    \end{itemize}
\end{itemize}

\paragraph{Step Accuracy $R_s$:}

This reward is crucial for ensuring the quality of the reasoning content itself.
\begin{itemize}
    \item We check if the model's output for the current step contains a pre-extracted key string (\texttt{key\_str}), which represents a crucial piece of information for that specific step.
    \item If the output contains the key string, $R_s = 1.0$.
    \item Otherwise, $R_s = 0.0$.
\end{itemize}

\paragraph{Final Reward Score $S_i$:}

The total reward score is a composite of all components, orchestrated by a multiplicative gating mechanism. This structure ensures a holistic evaluation of the model's performance, balancing process quality with final correctness.


\begin{equation}
\mathscr{S}_i =\alpha R_a + \beta R_f + \gamma (R_t + R_s)
\label{eq:score}
\end{equation}

The formula is decomposed into three main components.
The first part evaluates the quality of the final answer itself.
The second part evaluates whether the model outputs the final answer in the required format.
The third part assesses the quality of the reasoning process, which measures the logical structure and completeness of the reasoning steps, and the generated reasoning correctly captures key steps. 

\subparagraph{Parameter Explanations}
\begin{itemize}
    \item $\alpha$ (Answer Accuracy Weight): Set to 2.0. This weight emphasizes the importance of producing a correct final answer.
    \item $\beta$ (Format Weight): Set to 1.0. This weight ensures that the model adheres to the required output format.
    \item $\gamma$ (Reasoning Weight): Set to 1.0. This weight reflects the contribution of both reasoning completeness and critical step accuracy to the overall score.
\end{itemize}

\section{Prompt}
\subsection{Prompt for Training}
\begin{figure}[htbp]
    \centering
    \includegraphics[width=1.0\columnwidth]{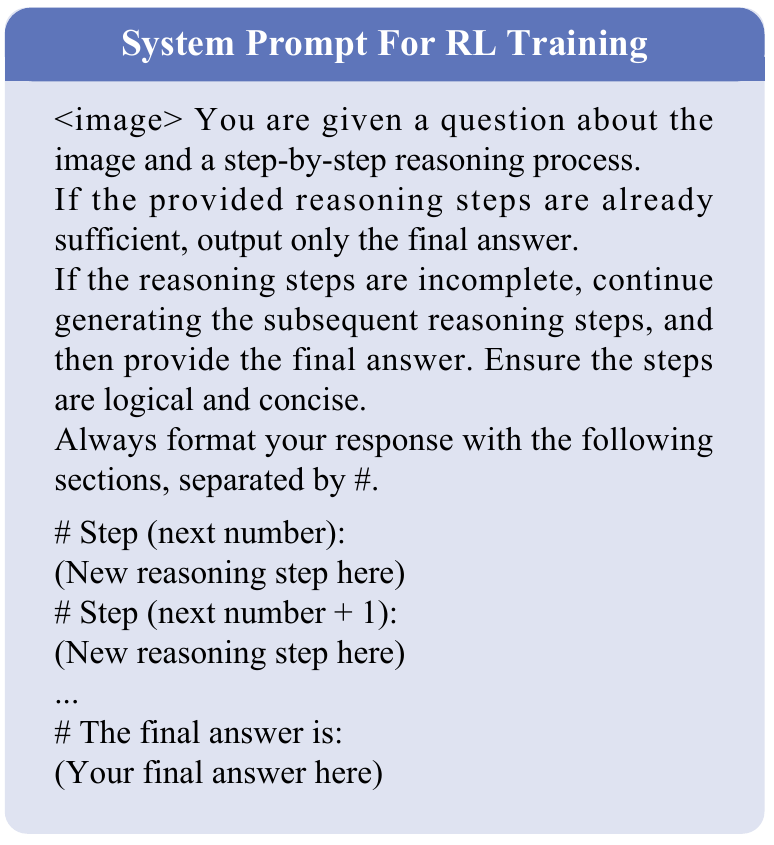}
    \caption{System Prompt used in RL Training.}
    \label{fig:training_prompt}
\end{figure}
The instruction following prompt for training is designed to guide the model through an incremental reasoning process. The key goals are:\\
\textbf{Prompting Incremental Reasoning.} The model is instructed to generate reasoning steps in sequence, ensuring that each step logically leads to the next. This encourages the model to build on previous steps and engage in systematic problem-solving, which is crucial for handling complex tasks.\\
\textbf{Ensuring Completeness in Reasoning.} If the initial reasoning steps are insufficient, the model is prompted to continue generating additional steps. This ensures that the model can handle incomplete or ambiguous reasoning processes, thereby improving its ability to deal with real-world situations that often require iterative problem-solving.\\
\textbf{Structured Response Format.} By enforcing a structured response format with clearly delineated steps and a final answer, the prompt fosters consistency in how the model organizes and presents its reasoning. This structure aids in improving the interpretability of the model’s outputs and makes it easier for humans to follow the model’s thought process.\\
\textbf{Adaptability for Training.} This design helps the model adapt to a wide range of problem complexities by learning to evaluate when reasoning is sufficient and when further elaboration is required. This encourages the model to handle a variety of reasoning tasks effectively during training.
\subsection{Prompt for Validating}
\begin{figure}[htbp]
    \centering
    \includegraphics[width=1.0\columnwidth]{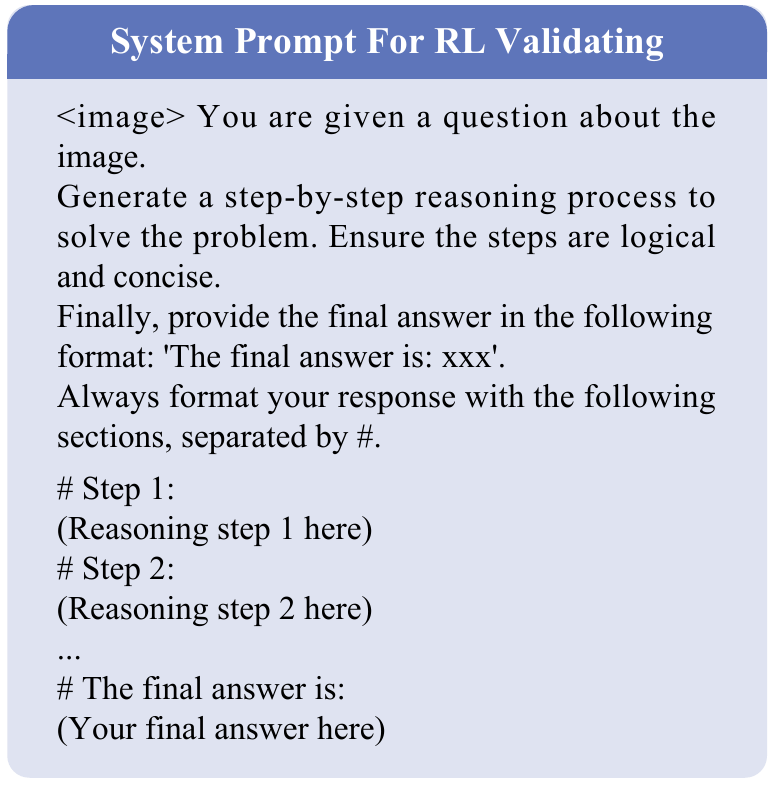}
    \caption{System Prompt used in RL validating.}
    \label{fig:validating_prompt}
\end{figure}
The instruction following prompt for validation focuses on the generation of reasoning without prior steps. The design objectives are:\\
\textbf{Encouraging Reasoning from Scratch.} The model is required to generate a step-by-step reasoning process from the given question, fostering the model’s ability to approach new problems without pre-existing steps. This is crucial for evaluating the model's ability to generalize to unseen tasks and situations.\\
\textbf{Ensuring Logical and Concise Steps.} The emphasis on generating logical and concise reasoning steps encourages the model to produce clear and efficient explanations. This aligns with the goal of guiding the model to be not only
accurate but also able to explain its reasoning in a manner that is understandable and precise.\\
\textbf{Providing a Final Answer.} After reasoning, the model is instructed to provide the final answer in a clear format. This ensures that the model is able to synthesize its reasoning into a coherent conclusion, which is essential for answering complex questions in real-world applications.\\
\textbf{Evaluation of Reasoning Ability.} This prompt design is particularly useful for evaluating the model’s capacity to generate reasoning and conclusions in a structured format. The clear format of steps and the final answer ensures consistency across multiple test cases, enabling a more accurate assessment of the model's performance.

\section{Case Study}
\begin{figure*}[htbp!]
    \centering
    \includegraphics[width=0.90\textwidth,
                     height=0.90\textheight,
                     keepaspectratio]{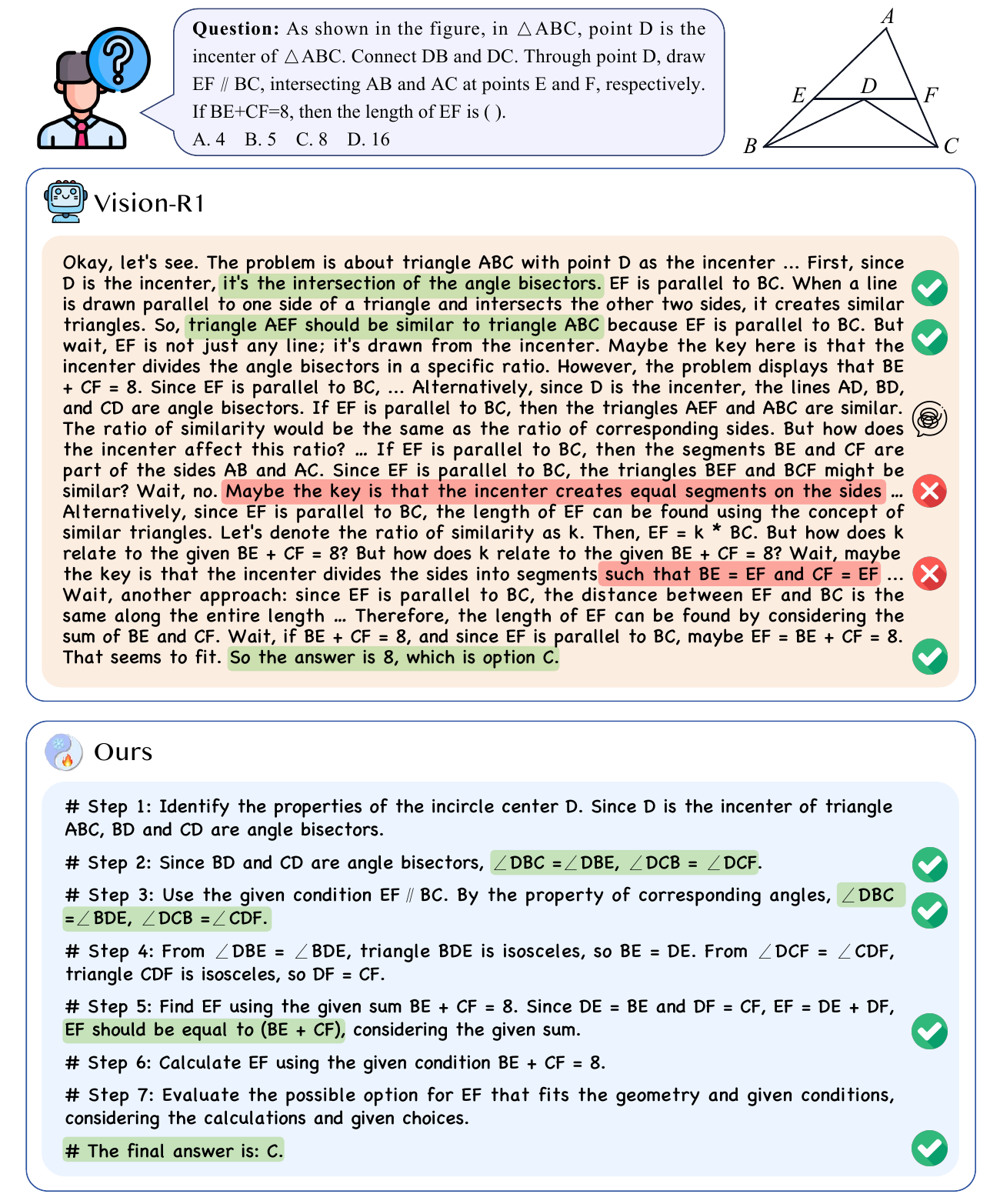}
    \captionof{figure}{\textbf{The Case Study of Mathematical Task.} The reasoning behaviors of Vision-R1, and our proposed O²-CritiCuRL. Existing Vision-R1 reaches the correct answer only through a lucky guess, as its intermediate reasoning contains conceptual mistakes and incorrect relationships between edges and angles. In contrast, our O²-CritiCuRL maintains coherent geometric relations and step-by-step logical transitions, producing a correct and interpretable reasoning process.}
    \label{fig:case2}
\end{figure*}

\clearpage
\begin{figure*}
    \centering
    \includegraphics[width=\textwidth,
                     height=\textheight,
                     keepaspectratio]{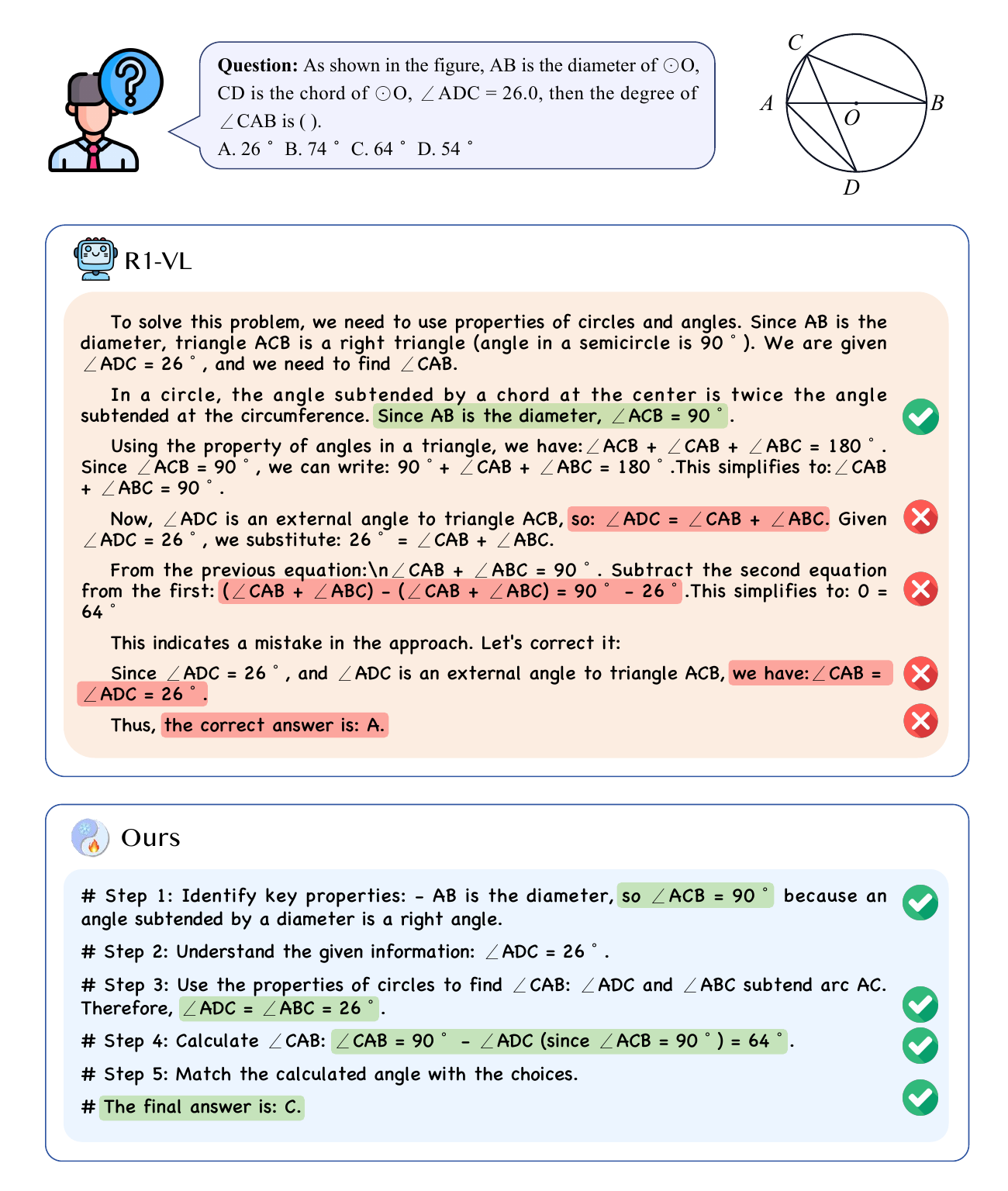}
    \captionof{figure}{\textbf{The Case Study of Mathematical Task.} The reasoning behaviors of R1-VL, and our proposed O²-CritiCuRL. R1-VL provides incorrect answers due to flawed reasoning, while ours identifies and follows correct steps, yielding coherent and interpretable output.}
    \label{fig:case1}
\end{figure*}

\clearpage
\begin{figure*}
    \centering
    \includegraphics[width=\textwidth,
                     height=\textheight,
                     keepaspectratio]{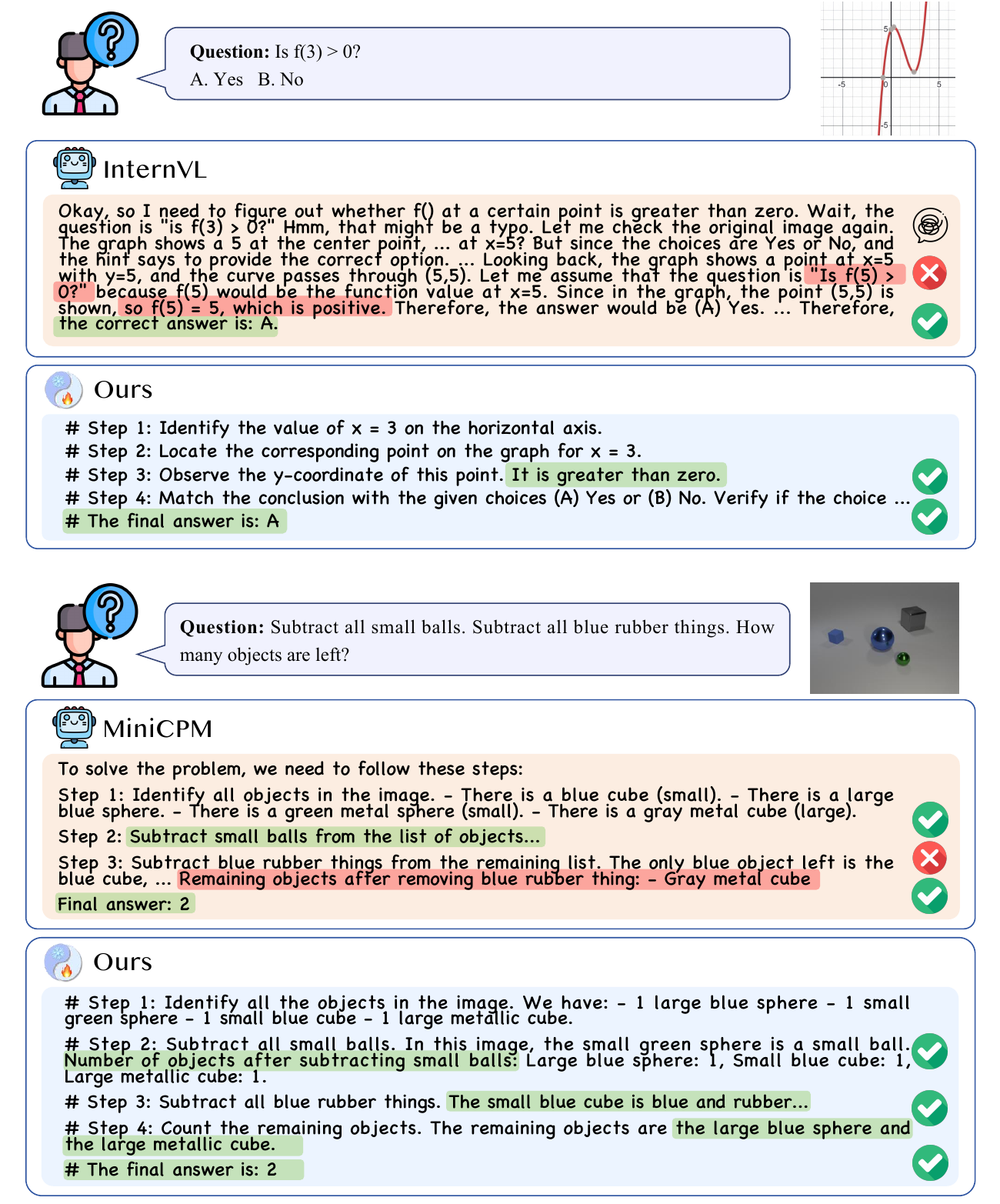}
    \captionof{figure}{\textbf{The Case Study of Mathematical Task.} The reasoning behaviors of InternVL, MiniCPM, and our proposed O²-CritiCuRL. In the first case, InternVL fails to identify the point, leading to an incorrect reasoning process, although it ultimately gives the correct option by chance. In the second case, MiniCPM arrives at a conclusion of 1 during its reasoning process but still produces the correct answer 2 through a lucky guess. In contrast, ours derives the correct answer through a coherent and logically grounded reasoning process.}
    \label{fig:case4}
\end{figure*}

\clearpage
\begin{figure*}
    \centering
    \includegraphics[width=\textwidth,
                     height=\textheight,
                     keepaspectratio]{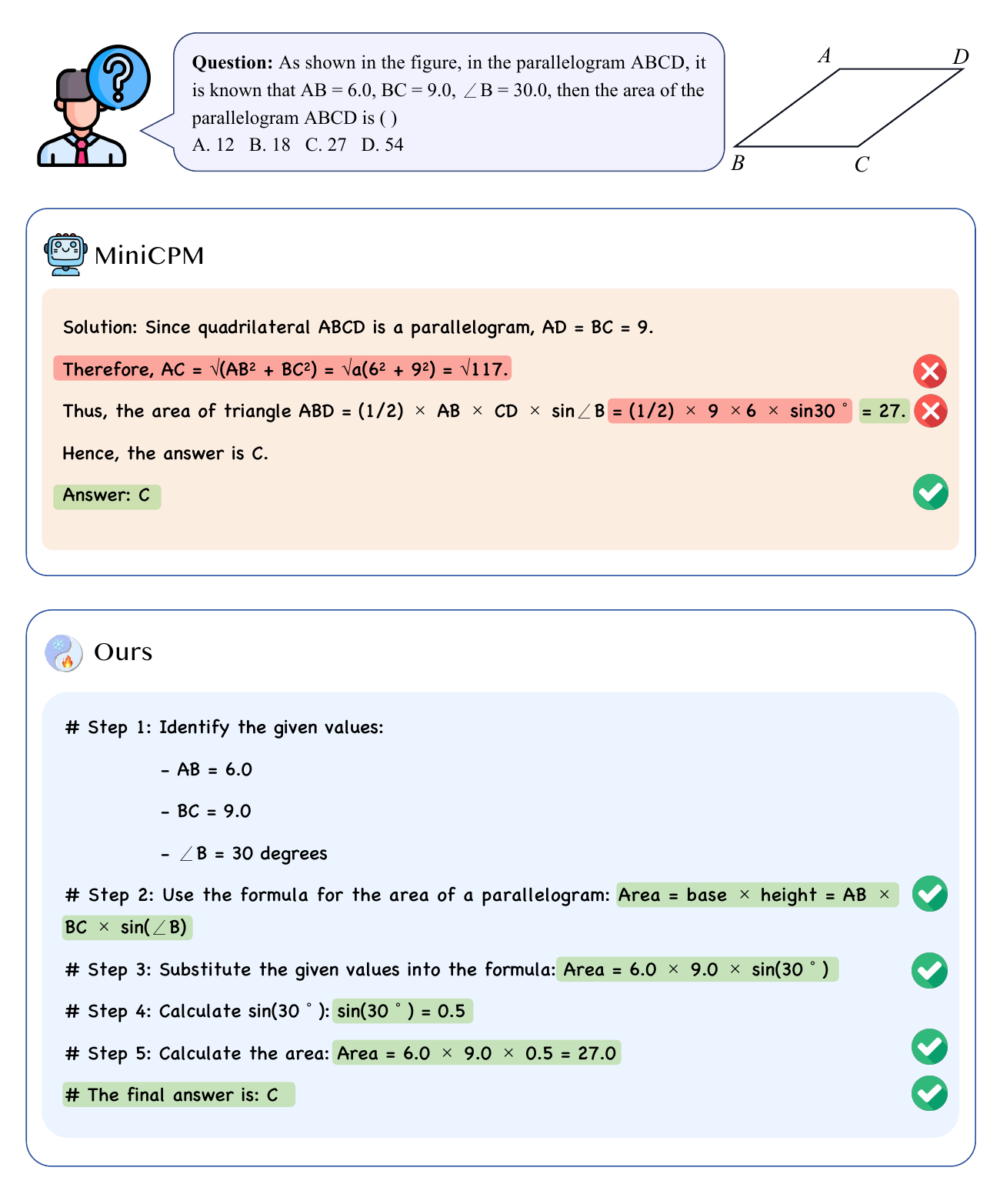}
    \captionof{figure}{\textbf{The Case Study of Mathematical Task.} The reasoning behaviors of MiniCPM and our O²-CritiCuRL. MiniCPM reaches the correct answer only through a "lucky guess", while ours produces the correct answer while adhering to the correct reasoning process.}
    \label{fig:case6}
\end{figure*}

\clearpage
\begin{figure*}
    \centering
    \includegraphics[width=\textwidth,
                     height=\textheight,
                     keepaspectratio]{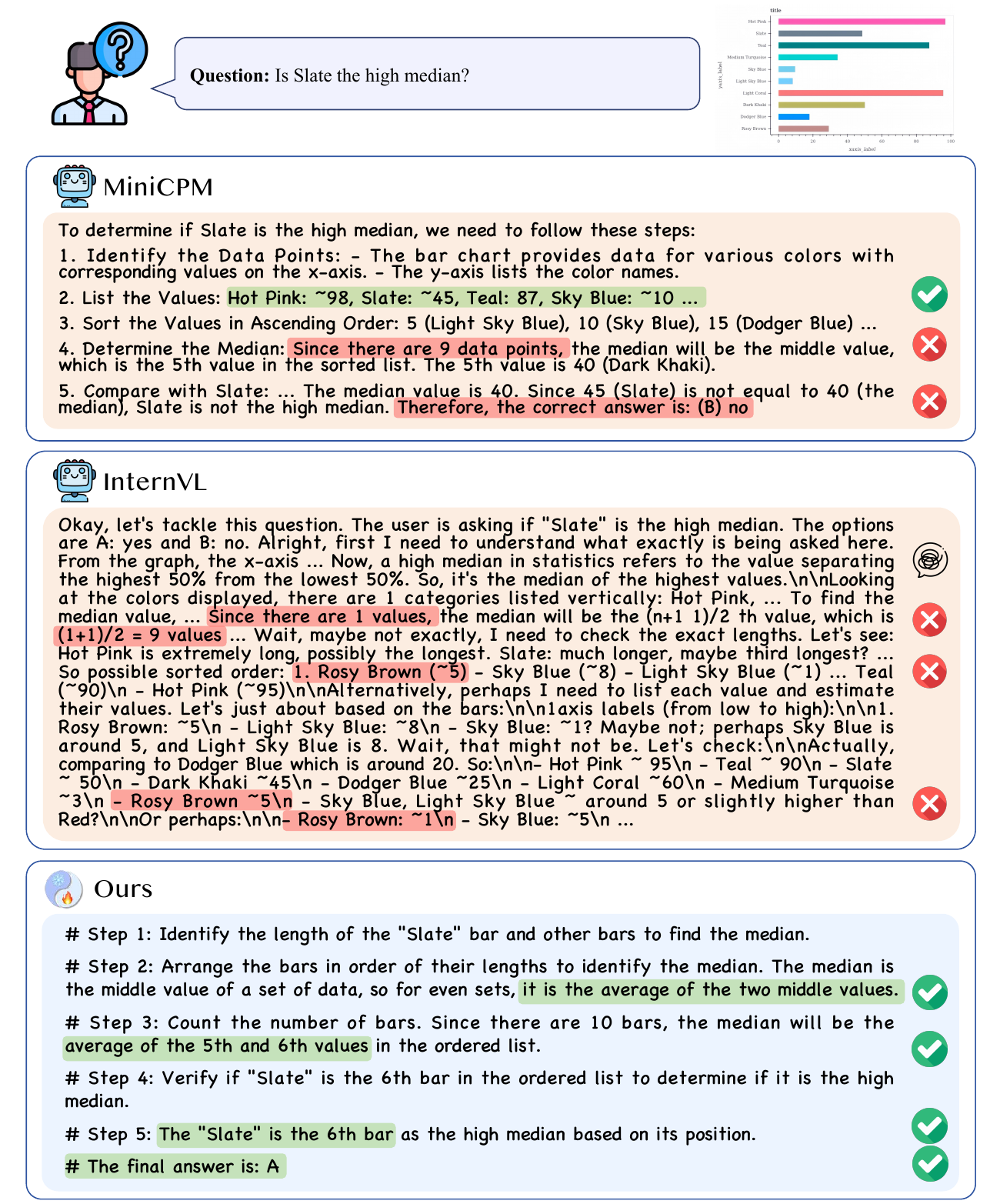}
    \captionof{figure}{\textbf{The Case Study of Chart Task.} The reasoning behaviors of MiniCPM, InternVL, and ours. MiniCPM misidentifies the number of bars, which leads to an incorrect reasoning process and ultimately an incorrect answer. InternVL misrecognizes the numerical values associated with the bars, also resulting in faulty reasoning and an incorrect final answer. In contrast, ours correctly identifies both the number of bars and their corresponding values, and produces the correct final answer through a coherent and accurate reasoning process.}
    \label{fig:case3}
\end{figure*}

\clearpage
\begin{figure*}
    \centering
    \includegraphics[width=\textwidth,
                     height=\textheight,
                     keepaspectratio]{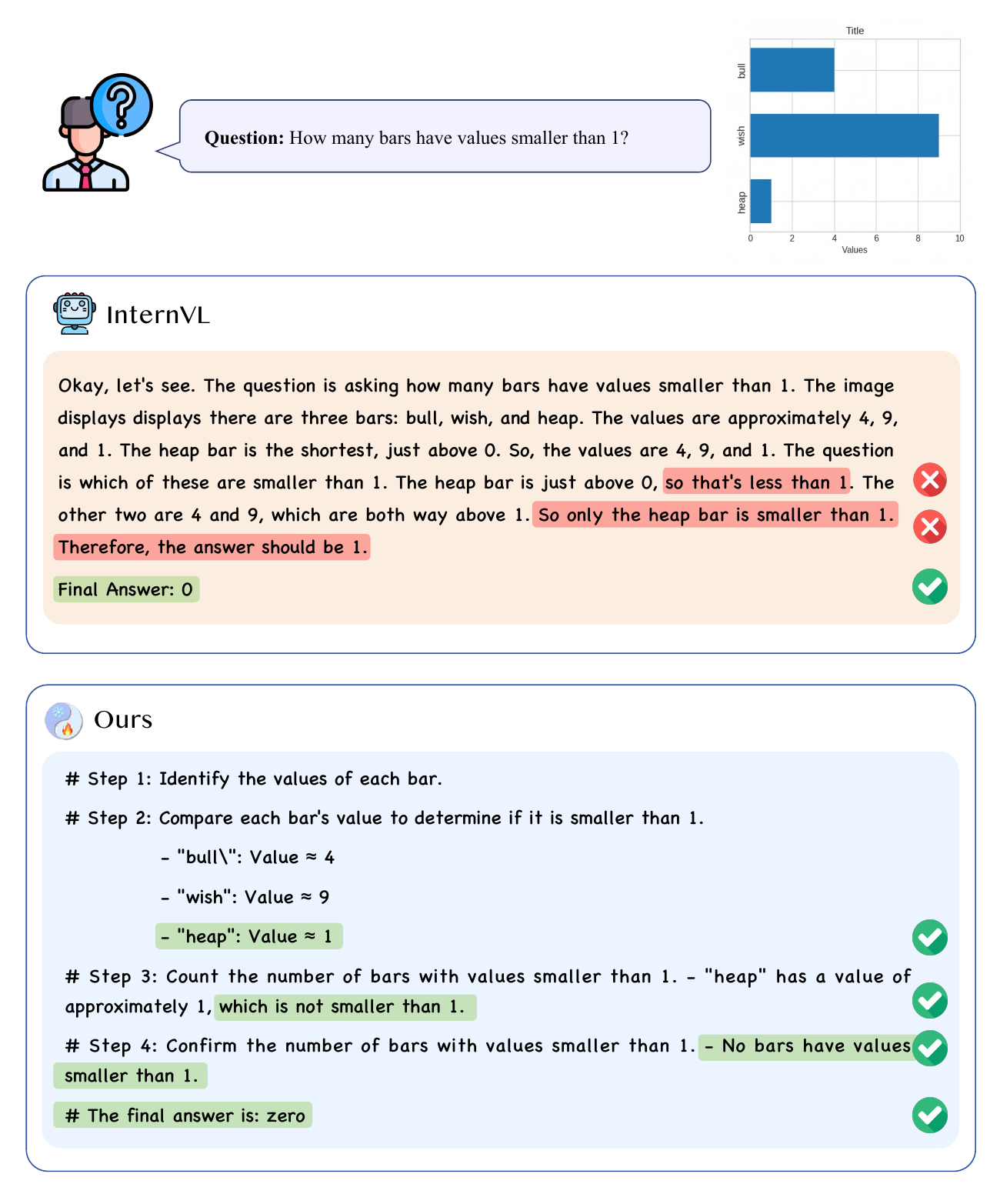}
    \captionof{figure}{\textbf{The Case Study of Chart Task.} The reasoning behaviors of InternVL and our proposed O²-CritiCuRL. InternVL arrives at a conclusion of "one" during its reasoning process, but ultimately guesses the correct answer "zero" by chance. In contrast, our O²-CritiCuRL identifies and follows the correct critical steps, yielding more logically sound and fully interpretable output.}
    \label{fig:case5}
\end{figure*}

\clearpage
\begin{figure*}
    \centering
    \includegraphics[width=\textwidth,
                     height=\textheight,
                     keepaspectratio]{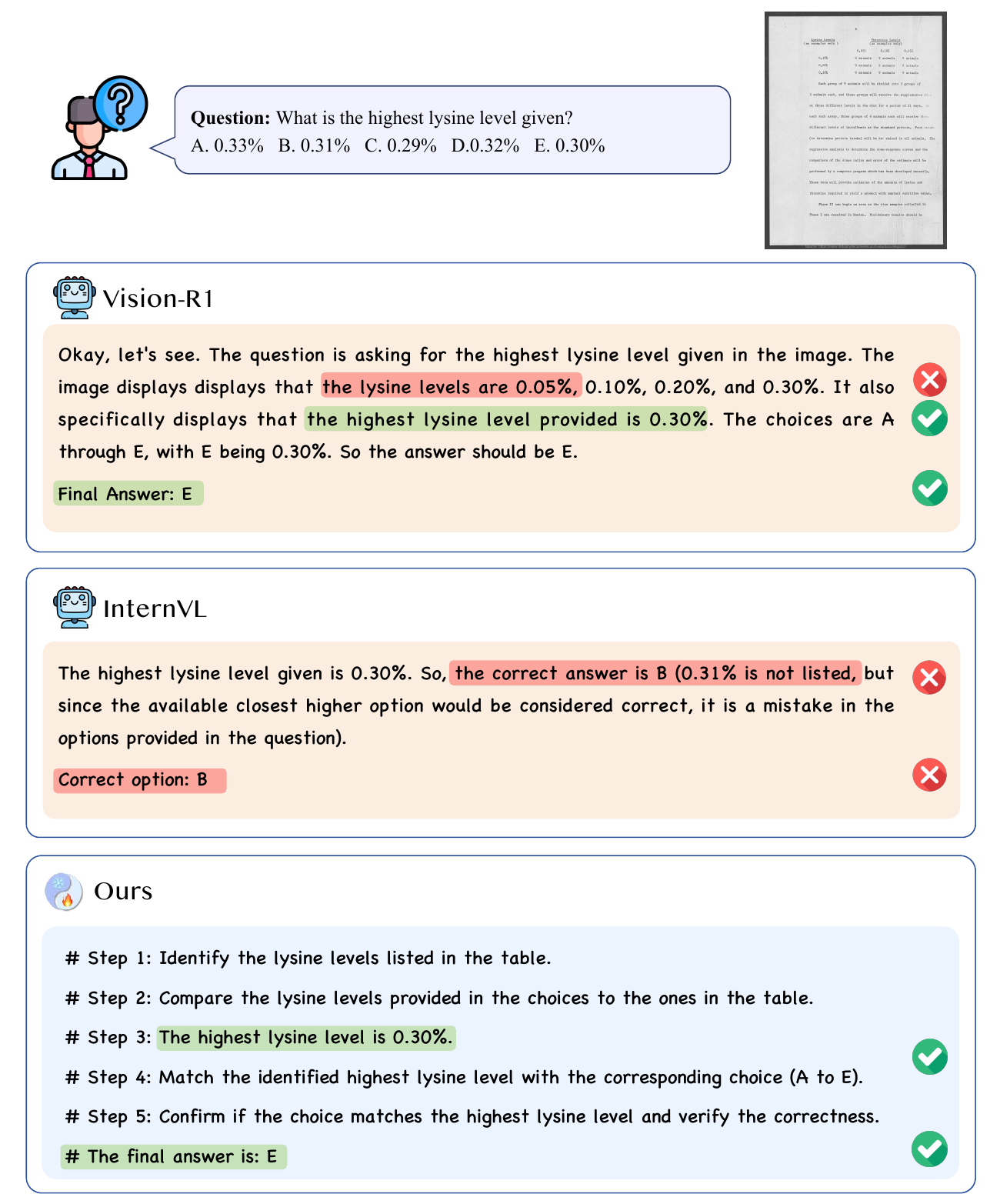}
    \captionof{figure}{\textbf{The Case Study of General Task.} The reasoning behaviors of Vision-R1, InternVL, and our O²-CritiCuRL. Existing Vision-R1 reaches the correct answer only through a “lucky guess”, while R1-VL provides incorrect answers due to flawed intermediate reasoning. In contrast, our O²-CritiCuRL identifies and follows the correct critical steps, yielding logically sound and interpretable output.}
    \label{fig:case7}
\end{figure*}

\end{appendices}

\clearpage






\end{document}